\documentclass{article}
\PassOptionsToPackage{table}{xcolor}
\usepackage[margin=1in]{geometry}
\usepackage{titling}
\usepackage{times}
\usepackage{natbib}
\setcitestyle{authoryear,round,citesep={;},aysep={,},yysep={;}}

\usepackage{amsmath,amsfonts,bm}

\def\eqref#1{equation~\ref{#1}}

\def\1{\bm{1}}

\DeclareMathAlphabet{\mathsfit}{\encodingdefault}{\sfdefault}{m}{sl}
\SetMathAlphabet{\mathsfit}{bold}{\encodingdefault}{\sfdefault}{bx}{n}

\usepackage{url}

\title{Architect-Ant: Editable Automatic Furnishing of Architectural Floor Plans}

\usepackage{placeins}
\usepackage{float}
\usepackage{graphicx}
\usepackage{booktabs}
\usepackage{tabularx}
\usepackage{xcolor}
\usepackage{tcolorbox}
\tcbuselibrary{breakable,skins}
\newtcolorbox{comparisonprompt}[1]{
  enhanced,breakable,colback=black!3,colframe=black!20,
  colbacktitle=black!8,coltitle=black,fonttitle=\bfseries\small,
  fontupper=\small,title={#1},boxrule=0.4pt,arc=2pt,
  left=8pt,right=8pt,top=6pt,bottom=6pt,
  before skip=8pt,after skip=8pt
}
\usepackage{pifont}
\usepackage{hyperref}
\usepackage{orcidlink}
\hypersetup{
  hidelinks,
  pdftitle={Architect-Ant: Editable Automatic Furnishing of Architectural Floor Plans},
  pdfauthor={Fedor Rodionov, Aleksandar Cveji\'c, Michael Birsak, John Femiani, Peter Wonka}
}

\author{%
  Fedor Rodionov\textsuperscript{1}\,\orcidlink{0009-0009-4306-0154}\quad
  Aleksandar Cveji\'c\textsuperscript{1}\,\orcidlink{0009-0005-4414-4457}\quad
  Michael Birsak\textsuperscript{1}\,\orcidlink{0000-0001-6375-8124}\\[0.4em]
  John Femiani\textsuperscript{2}\,\orcidlink{0000-0002-0924-6686}\quad
  Peter Wonka\textsuperscript{1}\,\orcidlink{0000-0003-0627-9746}\\[0.8em]
  \small\textsuperscript{1}King Abdullah University of Science and Technology (KAUST), Saudi Arabia\\
  \small\textsuperscript{2}Miami University, United States of America
}
\date{}

\newcommand{\cmark}{\textcolor{green!60!black}{\ding{51}}}
\newcommand{\xmark}{\textcolor{red}{\ding{55}}}

\newcommand{\best}[1]{\cellcolor{green!35}#1}
\newcommand{\second}[1]{\cellcolor{green!15}#1}

\begin{document}
\maketitle
\raggedbottom

\begin{figure}[H]
  \centering
  \includegraphics[width=\linewidth]{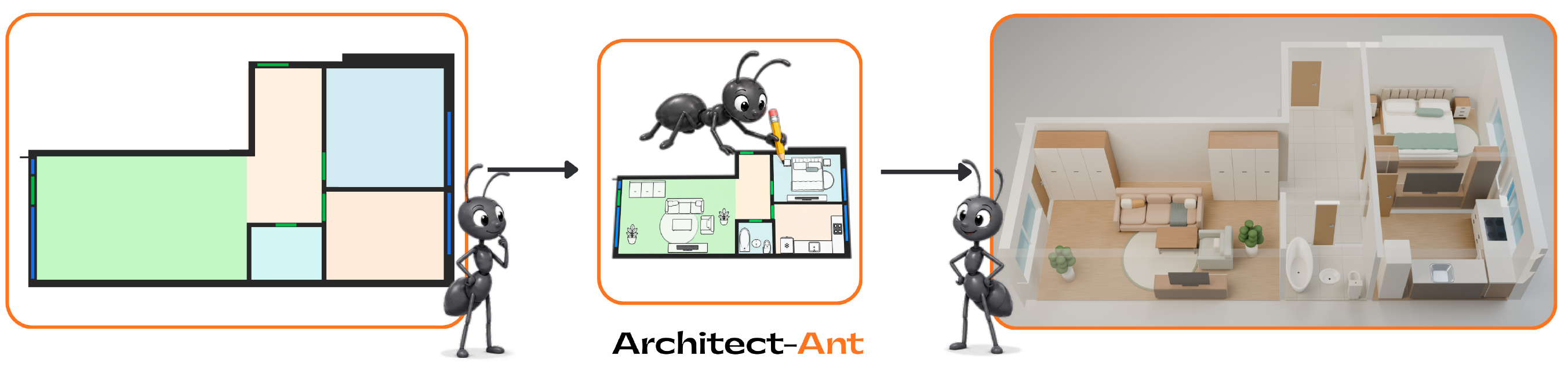}
  \caption{Architect-Ant transforms an empty structured floor plan into a furnished 3D scene.}
  \label{fig:teaser}
\end{figure}

\begin{abstract}
Furnished floor plans support real-estate visualization, interior design, and architectural workflows, yet automatic furnishing remains challenged by limited real-world data and the need to satisfy interacting geometric and functional constraints. We ask whether professional furnishing knowledge can be learned from real floor plans using a pretrained model, enabling direct constraint-aware layout generation without relying on costly iterative agentic inference. We introduce AntPlan, a curated dataset of 505 real professional architectural floor plans with dense furniture annotations spanning 92 object classes and ten residential room categories, and Architect-Ant, a framework for generating furniture layouts. Architect-Ant represents layouts with an editable coordinate-based DSL and first learns professional furnishing patterns through supervised fine-tuning. It is then optimized with GRPO using a Layout Rule Score (LRS) that aggregates geometric and functional constraints derived from professional plans, providing outcome-level supervision without prescribed reasoning traces. Experiments against diverse state-of-the-art baselines show that Architect-Ant combines low geometric violation rates with high functional completeness, while qualitative results more closely reflect real-world residential furnishing patterns. The resulting layouts remain object-level editable and can be converted into 3D scenes.
\end{abstract}

\section{Introduction}
\label{sec:intro}

Furnished floor plans are a practical starting point for interior design and for constructing 3D scenes used in architectural visualization, robotics simulation, and game development. Furniture drawn on a 2D plan conveys the scale and purpose of a room, how its occupants might use the space, and where they can move. For interior designers, furnished floor plans support the exploration and communication of alternative furniture arrangements and spatial configurations. For downstream applications, they provide the layout information from which a furnished scene can be built.

Producing such layouts manually requires precision and meticulous attention to constraints that extend well beyond a list of objects. A bed must fit within the room, but its placement also affects access to storage, nearby furniture, and circulation. A kitchen requires appliances and work surfaces to form usable arrangements, while doors, windows, and working areas must remain accessible. Many of these requirements are implicit in professional practice and interact with one another: moving one object can improve one relationship while breaking another. Automatic furnishing could increase productivity and support a wider range of alternatives while preserving these geometric and functional requirements and allowing designers to edit the result.

Good furniture placement draws on three complementary sources of knowledge. \textbf{Layout data} provides examples of object choices and spatial arrangements~\citep{Paschalidou2021ATISS,Tang2024DiffuScene,Lin2024InstructScene}; \textbf{explicit spatial constraints} make geometric and functional requirements computable~\citep{Yu2011MakeItHome,Merrell2011InteractiveFurniture}; and \textbf{pretrained models} contribute broad semantic and contextual knowledge~\citep{Feng2023LayoutGPT,yang_holodeck_2024,sun2024layoutvlm,Pfaff2026SceneSmithAG}. Each source is incomplete on its own: datasets are limited by their coverage, rules capture only encoded relationships, and pretrained models do not guarantee precise spatial placement. Recent agentic approaches can improve layouts through iterative reasoning and correction, but each refinement round increases inference cost and time. We ask whether these three sources can instead be combined during training so that a pretrained model learns to generate constraint-aware layouts directly.

A further challenge is obtaining suitable supervision. Public 3D scene datasets can be projected into 2D layouts, but such projections contain only the architectural detail represented in the source and do not necessarily reflect the conventions and construction elements found in professional floor plans. We therefore introduce \textbf{AntPlan}, a curated dataset of 505 real professional architectural floor plans with room-level structural annotations and dense furniture annotations spanning 92 object classes and ten residential room categories.

Building on AntPlan, we present \textbf{Architect-Ant}, a framework that combines professional layout examples, pretrained model knowledge, and explicit spatial feedback. It represents layouts with a compact coordinate-based domain-specific language (DSL) specifying object categories, positions, and dimensions, preserving object-level editability. We first adapt a pretrained model through supervised fine-tuning on detector-assisted, manually reviewed layouts. We then apply Dr.~GRPO-style reinforcement learning~\citep{Liu2025UnderstandingRT} using a room-aware \textbf{Layout Rule Score (LRS)} that aggregates geometric and functional constraints derived from professional layouts. Following outcome supervision~\citep{DeepSeekAI2025DeepSeekR1IR}, rewards are applied to the final layout rather than to prescribed reasoning traces, allowing the model to develop its own strategy for satisfying spatial requirements.

In experiments across four room types, Architect-Ant achieves the highest functional-completeness and overall layout-rule scores among the compared methods while maintaining low geometric violation rates, and generates layouts substantially faster than iterative agentic approaches.

\begin{table}[!t]
\setlength{\abovecaptionskip}{0pt}
\caption{Comparison of floor-plan datasets}
\label{tab:dataset_comparison}
\begin{center}

\scriptsize
\setlength{\tabcolsep}{4pt}
\renewcommand{\arraystretch}{1.12}

\begin{tabular*}{\columnwidth}{
@{\extracolsep{\fill}}
l
c
l
l
c
c
c
c
@{}
}
\hline

\textbf{Dataset} &
\textbf{\#} &
\textbf{Style} &
\textbf{Structural elements} &
\shortstack{\textbf{Room}\\\textbf{cls.}} &
\shortstack{\textbf{Avg.}\\\textbf{\# rooms}} &
\shortstack{\textbf{Furn.}\\\textbf{cls.}} &
\shortstack{\textbf{Avg.}\\\textbf{\# objects}} \\

\hline

MSD &
5,372 &
Synth., simple boxes &
\cmark\ wall/column, door, window &
9 &
30.78 &
\xmark &
\xmark \\

ResPlan &
17,000 &
Synth., simple boxes &
\cmark\ wall, door, window, front door &
7 &
8.22 &
\xmark &
\xmark \\

ZInD &
2,737 &
Synth., from 3D recon. &
\cmark\ door, window &
Free-text &
14.28 &
\xmark &
\xmark \\

FloorPlanCAD &
2,845$^{*}$ &
CAD drawings, tiled &
\cmark\ one class: structural &
\xmark &
\xmark &
\cmark\ 30 &
16.73 \\

SESYD &
10$^{**}$ &
Synthetic &
\cmark\ door, window &
\xmark &
\xmark &
\cmark\ 12 &
17.07 \\

CubiCasa5K &
5,000 &
Real, prof. arch. &
\cmark\ wall, door, window, railing &
10 &
12.18 &
\cmark\ 10$^{***}$ &
27.20 \\

\hline

\textbf{AntPlan} &
505 &
Real, prof. arch. &
\cmark\ wall, door, window, railing &
10 &
8.90 &
\cmark\ 92 &
53.01 \\

\hline
\end{tabular*}

\vspace{1mm}

\begin{minipage}{\columnwidth}
\scriptsize
$^{*}$FloorPlanCAD contains 2,845 source CAD drawings, distributed as
15,663 tiles/crops.\\
$^{**}$SESYD contains 10 unique layouts with 100 variations each. \\
$^{***}$CubiCasa5K furniture classes are coarse: appliances are not
distinguished, and some fixtures are represented only by partial symbols.
\end{minipage}

\end{center}
\end{table}

Our main contributions are:
\begingroup
\setlength{\topsep}{4pt}
\setlength{\partopsep}{0pt}
\setlength{\itemsep}{3pt}
\setlength{\parsep}{0pt}
\begin{itemize}
    \setlength{\parskip}{0pt}
    \raggedright
    \item \textbf{AntPlan}, a dataset of real professional architectural floor plans with dense furniture annotations across 92 object classes and ten residential room categories, substantially expanding furniture-class coverage over existing 2D floor-plan datasets.
    \item \textbf{Architect-Ant}, a framework that adapts pretrained model knowledge through supervised fine-tuning and rule-guided reinforcement learning to generate editable DSL layouts for downstream 3D scene construction.

    \item \textbf{A room-aware Layout Rule Score (LRS)} that converts geometric and functional constraints into layout-level reinforcement-learning rewards, enabling constraint-aware generation without prescribing intermediate reasoning steps.
\end{itemize}
\endgroup

\section{Related Work}
\label{sec:related}

\noindent\textbf{Floor plans and indoor scene datasets.}
RPLAN~\citep{Wu2019RPLAN}, Graph2Plan~\citep{Hu2020Graph2Plan}, House-GAN~\citep{Nauata2020HouseGAN}, and HouseDiffusion~\citep{Shabani2022HouseDiffusionVF} study architectural layout generation. Existing floor-plan datasets, including CubiCasa5K~\citep{kalervo2019cubicasa5k}, MSD~\citep{Engelenburg2024MSDAB}, ResPlan~\citep{Abouagour2025ResPlanAL}, ZInD~\citep{Cruz2021ZInD}, FloorPlanCAD~\citep{Fan2021FloorPlanCAD}, and SESYD~\citep{Delalandre2011RecentCO}, provide structural, room, or object annotations; Table~\ref{tab:dataset_comparison} compares their coverage with AntPlan. Indoor scene datasets such as 3D-FRONT and 3D-FUTURE~\citep{Fu20203DFRONT3F,Fu20203DFUTURE}, Structured3D~\citep{Zheng2020Structured3D}, HSSD~\citep{khanna2023hssd}, and ProcTHOR~\citep{Deitke2022ProcTHOR} provide object-level 3D scenes from which 2D layouts can be derived. AntPlan instead targets dense, fine-grained furniture annotation directly on real professional architectural floor plans.

\noindent\textbf{Furniture arrangement and scene synthesis.}
Classical systems optimize geometric and ergonomic objectives~\citep{Merrell2011InteractiveFurniture,Yu2011MakeItHome}, while learned approaches include autoregressive ATISS~\citep{Paschalidou2021ATISS}, DiffuScene~\citep{Tang2024DiffuScene}, and graph-based InstructScene~\citep{Lin2024InstructScene}. These learned methods train on furnished 3D-FRONT scenes, whereas Architect-Ant learns from furniture annotations on real architectural drawings. Prior methods incorporate individual sources of additional knowledge: DiffuScene uses a training-time overlap penalty, InstructScene uses pretrained text features for instruction conditioning, and other work studies constraint-aware generation or differentiable rule losses~\citep{Para2020GenerativeLM,Leimer2022LayoutEnhancer}. Architect-Ant adapts a pretrained language model to professional floor-plan annotations, then uses room-aware geometric and functional rewards to improve its generated layouts.

\noindent\textbf{Language and vision-language planning.}
LayoutGPT~\citep{Feng2023LayoutGPT} uses language models for structured layout generation with retrieved examples; we evaluate its zero-shot configuration. Holodeck~\citep{yang_holodeck_2024}, I-Design~\citep{cCelen2024IDesignPL}, and LayoutVLM~\citep{sun2024layoutvlm} combine language or visual reasoning with scene construction and spatial optimization, while SceneSmith~\citep{Pfaff2026SceneSmithAG} uses a hierarchical agentic pipeline for simulation-ready scene generation. Table~\ref{tab:knowledge-sources} summarizes the knowledge sources used by the compared methods. Architect-Ant incorporates spatial feedback during training; at inference, it generates candidate layouts directly and ranks them with LRS, without a model-based critique-and-revision loop. Recent evaluation work such as SceneEval~\citep{Tam2025SceneEvalES} further motivates assessing explicit geometric and functional properties alongside visual quality.

\begin{table}[!t]
\setlength{\abovecaptionskip}{0pt}
\setlength{\belowcaptionskip}{5pt}
\centering
\caption{Sources of layout knowledge in the compared configurations. Architect-Ant combines professional layout examples, spatial feedback, and pretrained model knowledge.}
\label{tab:knowledge-sources}
\scriptsize
\setlength{\tabcolsep}{2.3pt}
\renewcommand{\arraystretch}{1.2}
\begin{tabular*}{\linewidth}{@{\extracolsep{\fill}}lcccccccc@{}}
\toprule
\textbf{Knowledge source} & ATISS & \shortstack{Instruct\\Scene} & \shortstack{Diffu\\Scene} & \shortstack{LayoutGPT} & LayoutVLM & Holodeck & SceneSmith & \textbf{Architect-Ant} \\
\midrule
Layout examples              & \cmark & \cmark & \cmark & \xmark & \xmark & \xmark & \xmark & \cmark \\
Spatial constraints           & \xmark & \xmark & \cmark & \xmark & \cmark & \cmark & \cmark & \cmark \\
Pretrained model knowledge    & \xmark & \cmark & \xmark & \cmark & \cmark & \cmark & \cmark & \cmark \\
\bottomrule
\end{tabular*}
\par\smallskip
\begin{minipage}{\linewidth}
\scriptsize
\end{minipage}
\end{table}

\noindent\textbf{Rule-guided post-training.}
Verifier-based learning uses execution or explicit criteria to provide training feedback~\citep{Le2022CodeRL,Dou2024StepCoder}. GRPO~\citep{Shao2024DeepSeekMath} optimizes policies using relative rewards among sampled responses without a learned value network, while DeepSeek-R1-Zero~\citep{DeepSeekAI2025DeepSeekR1IR} demonstrates rule-based outcome supervision without reasoning demonstrations. We apply this principle to furniture placement by rewarding the final layout with LRS rather than supervising intermediate reasoning. Related layout work includes OptiScene~\citep{Yang2025OptiScene}, which uses preference optimization; in contrast, Architect-Ant learns from scalar geometric and functional layout rewards.

\section{Architect-Ant}
\label{sec:method}

\begin{figure}[!t]
  \centering
  \includegraphics[width=\linewidth]{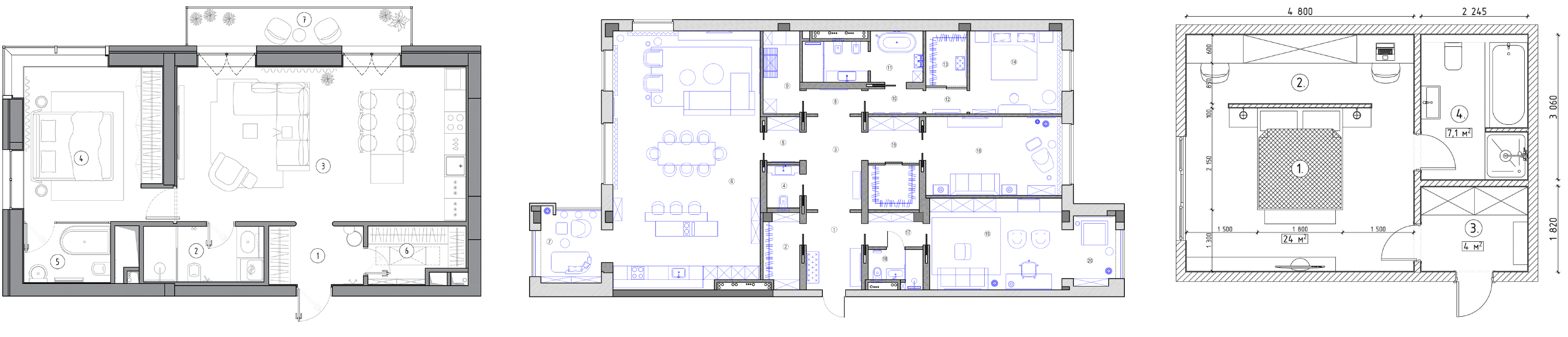}
  \caption{AntPlan floor plans with structural, room, and furniture annotations.}
  \label{fig:antplans}
\end{figure}

\begin{figure}[t]
  \centering
  \includegraphics[width=\linewidth]{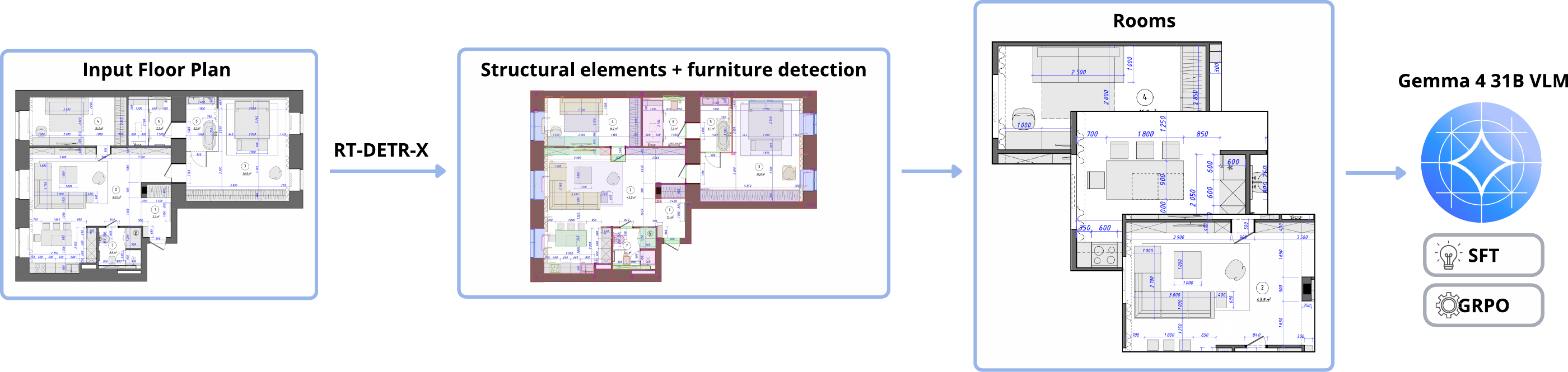}
  \caption{Data preparation and training. Detector-assisted annotations form room-level examples for supervised fine-tuning, followed by GRPO with final-layout rewards. A separate LoRA adapter is trained for each room category.}
  \label{fig:arch-ant-train}
\end{figure}

\begin{figure}[t]
    \centering
    \includegraphics[width=\linewidth]{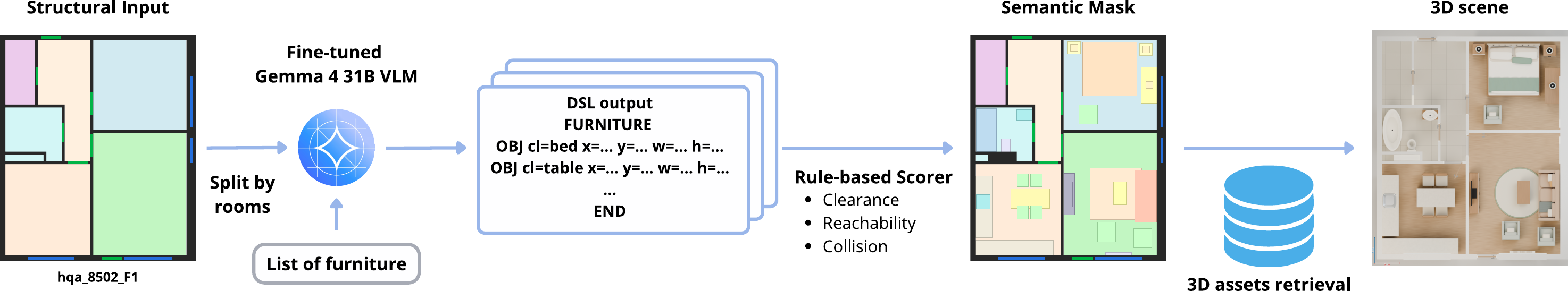}
    \caption{Inference and 3D scene construction. Room geometry, a structure-only image, and a retrieved furniture request condition the room adapter to generate a reasoning trace followed by an editable DSL layout. Layouts are scored and selected before 3D assets are retrieved and rendered.}
    \label{fig:arch-ant-gen}
\end{figure}

Architect-Ant combines three sources of layout knowledge: pretrained model knowledge, supervision from real architectural floor plans in AntPlan (Figure~\ref{fig:antplans}), and explicit spatial constraints used as reward supervision for GRPO. Given room geometry, a structure-only image, and a furniture request, it generates a coordinate-based DSL specifying object categories, positions, and dimensions. Figure~\ref{fig:arch-ant-train} summarizes supervised adaptation followed by GRPO with final-layout rewards, while Figure~\ref{fig:arch-ant-gen} shows inference and conversion of the generated layout into a 3D scene.

\subsection{AntPlan and room inputs}
\label{sec:method-dataset}

AntPlan contains 505 professional residential floor-plan drawings with structural, room, and furniture annotations across ten room categories. Walls, doors, windows, and railings are extracted using an RT-DETR-X detector~\citep{Lv2024RTDETRv2IB} trained on CubiCasa5K~\citep{kalervo2019cubicasa5k}, while room labels are assigned manually. A furniture detector is bootstrapped from a hand-labeled subset and applied to the remaining plans, followed by manual review and correction.

AntPlan provides 92 furniture classes, with an average of 53.01 annotated objects per plan, defined directly on professional 2D architectural drawings. Our experiments use cleaned bedroom, bathroom, kitchen, and living-room subsets containing 2,259 training rooms and 243 validation rooms. Within each room type, rooms from the same source floor plan are assigned exclusively to either the training or validation split. Further annotation and dataset statistics are provided in Appendix~\ref{app:dataset}.

Each room is represented by textual and visual inputs. The textual input specifies room dimensions and structural elements together with the requested furniture categories, counts, and approximate dimensions. The visual input is a structure-only raster that provides spatial context without exposing the target furniture layout. We train a separate adapter for each of the four evaluated room types. Appendix~\ref{app:visual-input} provides additional details on visual conditioning.

\subsection{Editable layout representation}
\label{sec:dsl}

A layout is represented as a sequence
$\mathcal{L}=\{(c_i,x_i,y_i,w_i,h_i)\}_{i=1}^{n}$,
where $c_i$ is the object class, $(x_i,y_i)$ is the top-left corner of its footprint, and $(w_i,h_i)$ are its extents along the room-local $x$ and $y$ axes.
Coordinates are expressed in meters relative to the room's top-left corner, with $x$ increasing rightward and $y$ downward. Appendix~\ref{app:representation} provides an example serialization.

This corner-based DSL provides a compact and editable representation while allowing containment, overlap, and clearance to be evaluated directly. The downstream conversion maps each footprint to its center and 3D asset transform. Asset dimensions and height come from the selected 3D model; orientation is inferred from furniture relationships and wall attachment rather than predicted by the DSL (Appendix~\ref{app:dsl-rendering}).

\subsection{Supervised adaptation}
\label{sec:method-sft}

We fine-tune Gemma-4-31B-it~\citep{gemmateam2026gemma4} using LoRA~\citep{Hu2021LoRALA} on the language model while keeping the vision encoder frozen. Training uses token-level cross-entropy on the furniture DSL and end-of-turn token, with prompt, image, and padding positions masked from the loss. The targets contain only the final layouts, without reasoning traces. SFT adapts the model to the furnishing domain and DSL representation; the subsequent GRPO stage optimizes layout quality using final-layout rewards while allowing intermediate reasoning. Training details are provided in Appendix~\ref{app:training-details}, and Appendix~\ref{app:reasoning} discusses experiments with supervised reasoning traces.

\subsection{Layout Rule Score and reward}
\label{sec:layout-score}

LRS is a deterministic room-aware score of geometric and functional constraint satisfaction. The evaluator parses the generated DSL and starts each layout at 10, deducting penalties for violated rule instances:
\begin{equation}
S(\mathcal{L};x)=10-\sum_k p_k(\mathcal{L},x),
\end{equation}
where $p_k\geq0$ is the penalty contributed by rule family $k$ across its applicable instances. Scores are unbounded below; a score of 10 indicates that no encoded rule was violated, rather than guaranteeing an ideal layout.

The evaluator implements 33 rule types spanning shared structural checks and room-specific functional relationships. Shared rules cover room containment, wall overlap and penetration, door and window obstruction, disallowed furniture overlap, inventory mismatch, and accessibility. Room policies add relationships such as appliance mounting, kitchen runs, and work zones, focal seating, window blocking, wall alignment, and fixture--furniture placement. Class-specific exemptions preserve valid arrangements, such as a rug beneath a bed or a sink mounted in a cabinet. The rules are instantiated over applicable objects, architectural elements, and object pairs, so the number of constraint checks grows with layout complexity; typical validation rooms require roughly 50--200 predicate evaluations.

Rule parameters, class sets, exemptions, and thresholds are calibrated against AntPlan annotations rather than requiring a generated layout to reproduce a particular reference. Room-specific policies activate additional constraints according to the furniture and functional relationships observed for that room type. Thresholds are selected from measured geometric distributions and checked against annotated layouts to avoid penalizing common valid arrangements. Appendix~\ref{app:scoring_details} provides the complete rule definitions, thresholds, and accessibility construction.

During GRPO, LRS provides the main outcome reward, augmented with terms for object-size deviation, gated same-class similarity to the annotated layout, and response validity. All these auxiliary terms are used only during training; inference ranks candidates by raw LRS. Answers with no parsed furniture receive a fixed negative training reward. Appendix~\ref{app:training-details} gives the exact reward terms.

LRS captures only the geometric and functional properties encoded by its rules; it does not measure aesthetics or guarantee full 3D usability. We therefore report individual geometric and functional diagnostics, object counts, and qualitative results alongside the aggregate score.

\subsection{GRPO with reasoning from layout rewards}
\label{sec:method-grpo}

We follow the R1-Zero principle of learning reasoning from outcome feedback~\citep{DeepSeekAI2025DeepSeekR1IR}, starting from the SFT checkpoint rather than the pretrained model. Each rollout contains a free-form reasoning block $r$ followed by a furniture layout $y$, where \(x\) denotes the room geometry and furniture request. The prompt encourages the model to consider alternative arrangements but does not prescribe the reasoning process. Reward is computed only from the parsed final layout, $R(r,y; x)=R(y; x),$
so intermediate reasoning receives no direct supervision.

For each training room, we sample 16 responses. Each update contains eight rooms, giving 128 on-policy rollouts, followed by a single policy-gradient step; rollouts are not reused. Advantages are computed by centering each reward by the mean of its room-level group, scaling by the pooled reward standard deviation across the batch, and clipping to $[-2,2]$. Following the motivation of Dr.~GRPO~\citep{Liu2025UnderstandingRT}, we avoid per-group variance normalization and normalize the policy loss by a fixed 1024-token constant rather than by individual response length. The resulting outcome advantage is applied to all generated tokens, including both reasoning and layout tokens.

We train for 15 updates with learning rate $2\times10^{-5}$ and KL coefficient $0.001$, regularizing against the frozen pretrained base model. The procedure uses no learned critic and performs one group-relative policy-gradient update per rollout batch. Full optimization, sampling, batching, and generation settings are provided in Appendix~\ref{app:training-details}.

\subsection{Candidate selection and 3D rendering}
\label{sec:inference}

For a new room shell, we retrieve same-type rooms from the AntPlan training set with similar dimensions and use their annotations to form furniture requests specifying object categories, counts, and approximate sizes. Architect-Ant generates layouts for each request and selects the layout with highest raw LRS, using geometric criteria only to break equal-score ties.

The selected DSL remains editable and can be converted into a furnished 3D scene. We retrieve assets with compatible categories and dimensions, fit them to the predicted footprints, and infer orientation from nearby walls and furniture relationships while retaining asset height. Appendix~\ref{app:dsl-rendering} provides rendering details.

\section{Experiments}
\label{sec:experiments}

We evaluate Architect-Ant in two settings. On held-out AntPlan rooms, we study the effects of supervised fine-tuning and reinforcement learning on layout quality. On released SceneSmith room shells, we compare the full Architect-Ant pipeline with agentic and learned baselines using geometric and functional criteria. Unless otherwise stated, the main comparison uses the reasoning-enabled GRPO model. Additional training-stage and direct-DSL results are provided in Appendix~\ref{app:reasoning}.

\subsection{Benchmark and comparison protocol}
\label{sec:protocol}

The comparison uses 109 released SceneSmith room shells: 31 bedrooms, 32 living rooms, 27 dining rooms, 13 kitchens, and six bathrooms. SceneSmith contributes its released furnished scenes, which incorporate its native iterative refinement; the remaining methods generate layouts for the corresponding room shells. Architect-Ant, Holodeck, LayoutGPT, and LayoutVLM use the same Gemma-4-31B backbone, replacing their original language models where applicable. ATISS, InstructScene, and DiffuScene cover the 90 bedroom, living-room, and dining-room shells; the latter two receive no target-shell geometry. For the Gemma-based methods, dining rooms use the kitchen setting; implementation details are provided in Appendix~\ref{app:comparison-details}.

For the regenerated methods, we evaluate four alternatives per room. Architect-Ant receives four furniture requests retrieved from same-type AntPlan training rooms with similar dimensions. Other regenerated baselines keep the same room specification and produce four alternatives through independent sampling or random seeds. One final layout is then selected per room.

The methods differ in how they process each alternative. Architect-Ant samples five layouts per furniture request, yielding 20 candidates per room ranked by LRS. Holodeck generates object choices and spatial constraints before searching for a feasible placement; LayoutGPT produces a layout in a single completion; LayoutVLM generates a constraint program and optimizes object placement; and ATISS, InstructScene, and DiffuScene each sample a layout from their learned models. SceneSmith uses repeated planner--designer--critic interactions with a scene-dependent number of revisions and a default budget of up to 1,000 turns per agent. All Architect-Ant candidate generation and selection are included in its reported inference time. Appendix~\ref{app:comparison-details} provides full protocol details.

From the SceneSmith scenes, we retain all assets labeled as furniture and, from its wall-mounted, manipuland, and ceiling-mounted classes, only objects whose categories are supported by our 53-class evaluation taxonomy.

\subsection{Metrics and comparison results}
\label{sec:metrics}

\begin{figure}[!p]
  \centering
  \includegraphics[width=\linewidth,height=0.9\textheight,keepaspectratio]{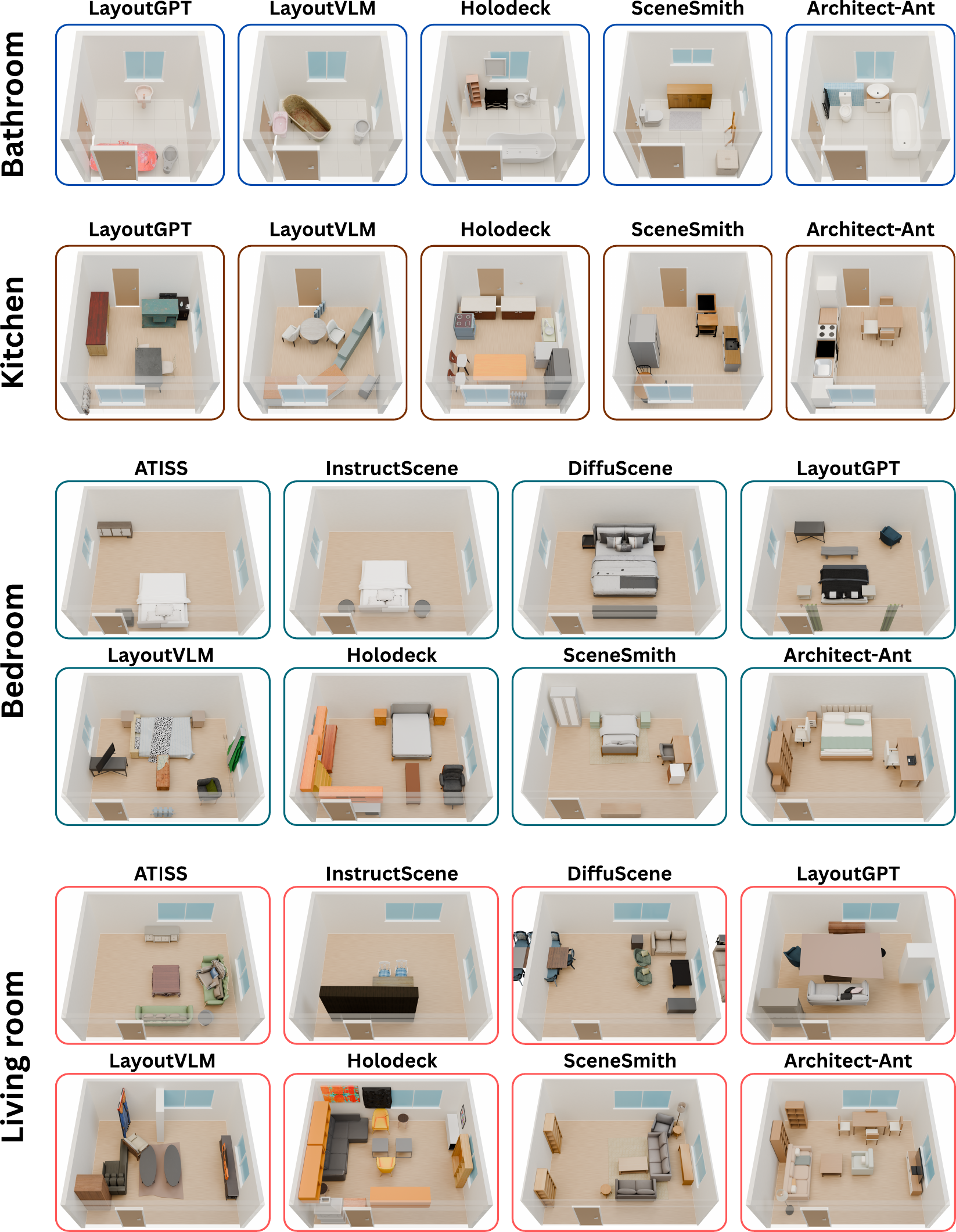}
    \caption{3D visualizations on shared room shells for bathrooms, kitchens, living rooms, and bedrooms. Learned 3D-FRONT baselines are shown only for supported room categories. The scenes provide qualitative comparisons; quantitative evaluation is performed on the layout geometry.}
  \label{fig:per-room-comparison}
\end{figure}

%

\begin{table}[!t]
\setlength{\abovecaptionskip}{0pt}
\caption{Comparison on SceneSmith room shells; values are room-level means. Dark/light green indicate the best/second-best geometric and functional results. COL+ and OOB are lower-is-better; $\uparrow$ indicates higher-is-better. Counts, coverage, and runtimes are not ranked.}
\label{tab:scene_generation_comparison}
\begin{center}

\scriptsize
\setlength{\tabcolsep}{0.4pt}
\renewcommand{\arraystretch}{1.15}
\newcommand{\comparisonheader}[1]{\begin{tabular}[c]{@{}c@{}}#1\end{tabular}}

\begin{tabular*}{\columnwidth}{
@{\extracolsep{\fill}}
l
c!{\hspace{3pt}}|!{\hspace{3pt}}
c
c
c
c!{\hspace{3pt}}|!{\hspace{3pt}}
c
c
c
c
c
c
@{}
}
\hline

\multicolumn{2}{c!{\hspace{3pt}}|!{\hspace{3pt}}}{} &
\multicolumn{4}{c!{\hspace{3pt}}|!{\hspace{3pt}}}{\textbf{Scene statistics}} &
\multicolumn{6}{c}{\textbf{Metrics and runtime}} \\
\hline
\comparisonheader{\textbf{Method}} &
\comparisonheader{\textbf{Base}\\\textbf{model}} &
\comparisonheader{\textbf{Scenes}} &
\comparisonheader{\textbf{\# Obj./}\\\textbf{room}} &
\comparisonheader{\textbf{\# Obj./}\\\textbf{10m$^2$}} &
\comparisonheader{\textbf{Cov.}\\\textbf{(\%)}} &
\comparisonheader{\textbf{COL+$\downarrow$}\\\textbf{(\%)}} &
\comparisonheader{\textbf{OOB$\downarrow$}\\\textbf{(\%)}} &
\comparisonheader{\textbf{Acc. pairs$\uparrow$}\\\textbf{(\%)}} &
\comparisonheader{\textbf{FUN+$\uparrow$}\\\textbf{(\%)}} &
\comparisonheader{\textbf{LRS$\uparrow$}} &
\comparisonheader{\textbf{Time$\downarrow$}\\\textbf{(s/scene)}} \\

\hline

ATISS &
Transformer &
90 &
6.5 &
3.10 &
29.5 &
24.1 &
18.1 &
95.2 &
84 &
1.50 &
20.2 \\

InstructScene &
Graph diff. &
90 &
7.2 &
3.15 &
26.3 &
13.7 &
20.9 &
86.9 &
63 &
$-2.82$ &
21.6 \\

DiffuScene &
Diffusion &
90 &
8.0 &
3.47 &
26.5 &
18.1 &
20.8 &
82.7 &
71 &
$-3.55$ &
74.3 \\

LayoutGPT &
Gemma-4 &
109 &
7.8 &
4.00 &
21.9 &
8.6 &
8.5 &
97.4 &
84 &
2.10 &
67.7 \\

LayoutVLM &
Gemma-4 &
109 &
8.2 &
4.22 &
28.1 &
12.7 &
16.0 &
94.0 &
86 &
6.29 &
272.0 \\

Holodeck &
Gemma-4 &
109 &
9.1 &
4.59 &
34.7 &
\best{0.0} &
\best{0.0} &
\best{99.5} &
78 &
7.84 &
432 \\

SceneSmith &
GPT-5.2 &
109 &
8.9 &
4.50 &
23.4 &
3.4 &
1.3 &
\second{99.4} &
\second{91} &
\second{8.44} &
1555 \\

\hline

\textbf{Architect-Ant} &
Gemma-4 &
109 &
10.8 &
5.45 &
29.1 &
\second{0.6} &
\second{0.6} &
97.2 &
\best{94} &
\best{8.80} &
95.4 \\

\hline
\end{tabular*}


\begin{minipage}{\columnwidth}
\scriptsize
\end{minipage}

\end{center}
\end{table}

Table~\ref{tab:scene_generation_comparison} reports room-level means. We separate descriptive scene statistics from geometric and functional quality metrics. \textbf{\# Obj./room}, \textbf{\# Obj./10m$^2$}, and \textbf{Coverage} characterize furnishing density. Coverage is the union of furniture footprints inside the room divided by floor area, excluding floor coverings and soft accessories; it measures occupancy rather than quality, so higher values are not necessarily better.

For geometric validity, \textbf{COL+} reports the fraction of objects involved in an invalid collision after category-aware filtering. Raw overlap is unsuitable because many furniture overlaps are intentional, such as chairs partly under tables, rugs beneath furniture, or inset fixtures. \textbf{OOB} measures the fraction of objects extending outside the room boundary. \textbf{Acc. pairs} measures the fraction of relevant furniture and door targets that remain mutually reachable. \textbf{FUN+} (functional completeness) is the percentage of required room functions supported by contained, reachable objects, such as sleeping, washing, or cooking. Appendix~\ref{app:functional-completeness} lists the required functions and eligibility checks.

\textbf{LRS} is the room-aware score defined in Section~\ref{sec:layout-score}. It combines geometric and functional penalties, including relationships not captured by collision or containment alone, and omits inventory-matching and training-only reward terms. Low COL+ or OOB therefore does not by itself imply a usable layout: objects may still block openings, violate functional relationships, or leave required functions unsatisfied. We interpret the metrics jointly with scene statistics and qualitative results. Appendix~\ref{app:comparison-details} defines the metrics; Appendix~\ref{app:reference-lrs} reports LRS for annotated AntPlan reference layouts.

Architect-Ant achieves the highest mean FUN+ and LRS while maintaining low collision and out-of-bounds rates. Within the evaluated furnishing categories, it also produces the highest mean number of furniture objects per room, indicating that geometric validity is not obtained through sparse layouts. Architect-Ant's accessible-pair rate (97.2\%) is within 2.3 points of the best method (Holodeck, 99.5\%). Qualitative comparisons in Figure~\ref{fig:per-room-comparison} reveal differences not fully captured by these metrics: Holodeck may obstruct doors or windows, SceneSmith can produce sparse arrangements or implausible orientations, and the learned baselines show visible object overlaps in several cases. Architect-Ant generally produces fuller arrangements while preserving openings, circulation, and functional relationships. Architect-Ant and all regenerated baselines are run on the same two NVIDIA H100 80GB GPUs. Reported times include candidate generation and scene construction; Architect-Ant's estimated total is 95.4\,s per room. SceneSmith's 1,555\,s entry is an object-count-normalized estimate from its full pipeline, not a measured rerun on the filtered objects. Appendix~\ref{app:comparison-details} gives the timing scopes and breakdowns.

\subsection{VLM-based visual preference}
\label{sec:visual-preference}

We complement geometric and functional diagnostics with a pairwise visual preference study on the 109 shared room shells. Claude Sonnet 5~\citep{anthropic2026sonnet5systemcard} and Kimi K3~\citep{moonshot2026kimik3} compare anonymous Architect-Ant and SceneSmith renders using the same prompt, images, and randomized A/B assignments. As shown in Table~\ref{tab:room-preference}, Sonnet prefers Architect-Ant in 72/109 rooms and Kimi in 74/109, with both showing the strongest preference on kitchen/dining shells; SceneSmith is preferred in four of the six bathrooms. LRS also favors Architect-Ant (8.80 vs.\ 8.44 overall), agreeing with the judges' aggregate preference while measuring geometric and functional constraints. Appendix~\ref{app:preference} provides the full protocol, prompts, whole-house comparison, and example pairs.

\FloatBarrier

\begin{table}[!t]
\setlength{\abovecaptionskip}{0pt}
\setlength{\belowcaptionskip}{5pt}
\centering\scriptsize
\caption{Pairwise visual preference on shared SceneSmith room shells, reported as Architect-Ant wins / ties / SceneSmith wins. The final row reports mean LRS for Architect-Ant / SceneSmith.}
\label{tab:room-preference}
\setlength{\tabcolsep}{2.5pt}
\begin{tabular*}{\linewidth}{@{\extracolsep{\fill}}lccccc@{}}
\toprule
Evaluator & Bathroom {\scriptsize($n=6$)} & Bedroom {\scriptsize($n=31$)} & Kitchen/dining {\scriptsize($n=40$)} & Living room {\scriptsize($n=32$)} & Overall {\scriptsize($n=109$)} \\
\midrule
Claude Sonnet 5 & 2 / 0 / \textbf{4} & \textbf{18} / 1 / 12 & \textbf{32} / 0 / 8 & \textbf{20} / 0 / 12 & \textbf{72} / 1 / 36 \\
Kimi K3 & 2 / 0 / \textbf{4} & \textbf{19} / 0 / 12 & \textbf{31} / 0 / 9 & \textbf{22} / 1 / 9 & \textbf{74} / 1 / 34 \\
\midrule
Mean LRS & 9.00 / \textbf{9.17} & \textbf{9.12} / 8.54 & \textbf{8.30} / 8.15 & \textbf{9.09} / 8.56 & \textbf{8.80} / 8.44 \\
\bottomrule
\end{tabular*}
\end{table}



\begin{table}[h]
\setlength{\belowcaptionskip}{5pt}
\caption{Validation LRS over valid layouts (mean $\pm$ CI half-width), with eight samples per room: (a) mean; (b) best-of-eight. Bold: row maximum.}
\label{tab:training_ablation}
\centering\scriptsize
\setlength{\tabcolsep}{1.5pt}
\renewcommand{\arraystretch}{1.15}
\begin{minipage}[t]{0.49\linewidth}
\centering
\begin{tabular*}{\linewidth}{@{\extracolsep{\fill}}lrrr@{}}
\multicolumn{4}{c}{\textbf{(a) Mean LRS}} \\
\toprule
\textbf{Room} & \multicolumn{1}{c}{\shortstack{\textbf{Base}\\DSL}} & \multicolumn{1}{c}{\shortstack{\textbf{SFT}\\DSL}} & \multicolumn{1}{c}{\shortstack{\textbf{SFT (DSL) +}\\\textbf{GRPO reasoning}}} \\
\midrule
Kitchen & $-11.71\,{\scriptstyle\pm\,3.62}$ & $-0.75\,{\scriptstyle\pm\,1.96}$ & $\mathbf{3.48}\,{\scriptstyle\pm\,1.82}$ \\
Bathroom & $-0.53\,{\scriptstyle\pm\,1.35}$ & $6.25\,{\scriptstyle\pm\,0.88}$ & $\mathbf{7.00}\,{\scriptstyle\pm\,0.74}$ \\
Living room & $2.29\,{\scriptstyle\pm\,1.50}$ & $5.41\,{\scriptstyle\pm\,0.93}$ & $\mathbf{6.22}\,{\scriptstyle\pm\,0.92}$ \\
Bedroom & $-1.51\,{\scriptstyle\pm\,1.58}$ & $5.71\,{\scriptstyle\pm\,0.96}$ & $\mathbf{6.02}\,{\scriptstyle\pm\,0.75}$ \\
\bottomrule
\end{tabular*}
\end{minipage}
\hfill
\begin{minipage}[t]{0.49\linewidth}
\centering
\begin{tabular*}{\linewidth}{@{\extracolsep{\fill}}lrrr@{}}
\multicolumn{4}{c}{\textbf{(b) Best-of-8 LRS}} \\
\toprule
\textbf{Room} & \multicolumn{1}{c}{\shortstack{\textbf{Base}\\DSL}} & \multicolumn{1}{c}{\shortstack{\textbf{SFT}\\DSL}} & \multicolumn{1}{c}{\shortstack{\textbf{SFT (DSL) +}\\\textbf{GRPO reasoning}}} \\
\midrule
Kitchen & $-2.37\,{\scriptstyle\pm\,2.90}$ & $5.73\,{\scriptstyle\pm\,1.21}$ & $\mathbf{5.80}\,{\scriptstyle\pm\,1.70}$ \\
Bathroom & $4.00\,{\scriptstyle\pm\,1.13}$ & $8.50\,{\scriptstyle\pm\,0.60}$ & $\mathbf{9.00}\,{\scriptstyle\pm\,0.43}$ \\
Living room & $6.54\,{\scriptstyle\pm\,1.02}$ & $8.77\,{\scriptstyle\pm\,0.55}$ & $\mathbf{8.95}\,{\scriptstyle\pm\,0.53}$ \\
Bedroom & $3.89\,{\scriptstyle\pm\,1.25}$ & $8.48\,{\scriptstyle\pm\,0.47}$ & $\mathbf{8.79}\,{\scriptstyle\pm\,0.46}$ \\
\bottomrule
\end{tabular*}
\end{minipage}
\end{table}

\subsection{Effect of SFT and GRPO}
\label{sec:ablation}

On the 243 held-out AntPlan rooms, we compare the base model with direct DSL output, five-epoch direct-DSL SFT, and SFT followed by reasoning-enabled GRPO. Each generates eight samples per room; only parseable, non-empty layouts are retained and scored by raw LRS. Table~\ref{tab:training_ablation} reports the retained-layout mean and best sample per room. Adding reasoning-enabled GRPO after SFT improves both mean and best-of-eight LRS in all four room types. Appendix~\ref{app:reasoning} compares this recipe with direct-DSL GRPO and with GRPO without SFT.

\section{Conclusion}
\label{sec:conclusion}

We presented Architect-Ant, a model for generating editable furniture layouts from architectural floor plans. Supervised fine-tuning on AntPlan teaches furnishing patterns, and GRPO improves layout quality using LRS rewards without supervised reasoning traces. The generated layouts can be evaluated geometrically and converted into furnished 3D scenes.

Architect-Ant achieves the highest mean functional-completeness (94\%) and LRS (8.80) among the compared methods, with collision and out-of-bounds rates of 0.6\% each. Both visual judges favor Architect-Ant over SceneSmith, preferring its layouts in 72 and 74 of the 109 paired rooms. These results show that direct layout generation can achieve strong geometric, functional, and visual performance without an iterative model-based critique loop. Future work can extend this framework to additional object and scene properties, richer geometry, and more detailed spatial relationships while preserving structured, editable generation.




\section*{Acknowledgments}
The work is supported by funding from King Abdullah University of Science and Technology (KAUST)---Center of Excellence for Generative AI, under award number 5940, and a gift from Google.

\bibliography{main}

@article{Lv2024RTDETRv2IB,
  title={{RT-DETRv2: Improved Baseline with Bag-of-Freebies for Real-Time Detection Transformer}},
  author={Wenyu Lv and Yian Zhao and Qinyao Chang and Kui Huang and Guanzhong Wang and Yi Liu},
  journal={ArXiv},
  year={2024},
  volume={abs/2407.17140},
  url={https://api.semanticscholar.org/CorpusID:271404151}
}

@article{Cruz2021ZInD,
  title={{Zillow Indoor Dataset: Annotated Floor Plans With 360° Panoramas and 3D Room Layouts}},
  author={Steve Dias da Cruz and Will Hutchcroft and Yuguang Li and Naji Khosravan and Ivaylo Boyadzhiev and Sing Bing Kang},
  journal={2021 IEEE/CVF Conference on Computer Vision and Pattern Recognition (CVPR)},
  year={2021},
  pages={2133-2143},
  url={https://api.semanticscholar.org/CorpusID:235694968}
}

@inproceedings{kalervo2019cubicasa5k,
  title={{CubiCasa5K: A Dataset and an Improved Multi-Task Model for Floorplan Image Analysis}},
  author={Ahti Kalervo and Juha Ylioinas and Markus H{\"a}iki{\"o} and Antti Karhu and Juho Kannala},
  booktitle={Scandinavian Conference on Image Analysis},
  pages={28--40},
  year={2019},
  organization={Springer},
  url={https://api.semanticscholar.org/CorpusID:102487507}
}

@inproceedings{Engelenburg2024MSDAB,
  title={{MSD: A Benchmark Dataset for Floor Plan Generation of Building Complexes}},
  author={Van Engelenburg, Casper and Mostafavi, Fatemeh and Kuhn, Emanuel and Jeon, Yuntae and Franzen, Michael and Standfest, Matthias and van Gemert, Jan and Khademi, Seyran},
  booktitle={European Conference on Computer Vision},
  pages={60--75},
  year={2024},
  organization={Springer},
  url={https://api.semanticscholar.org/CorpusID:271213468}
}

@article{Abouagour2025ResPlanAL,
  title={{ResPlan: A Large-Scale Vector-Graph Dataset of 17,000 Residential Floor Plans}},
  author={Abouagour, Mohamed and Garyfallidis, Eleftherios},
  journal={ArXiv},
  year={2025},
  volume={abs/2508.14006},
  url={https://api.semanticscholar.org/CorpusID:280686492}
}

@article{Fu20203DFRONT3F,
  title={{3D-FRONT: 3D Furnished Rooms with layOuts and semaNTics}},
  author={Huan Fu and Bowen Cai and Lin Gao and Ling-Xiao Zhang and Cao Li and Zengqi Xun and Chengyue Sun and Yiyun Fei and Yu-qiong Zheng and Ying Li and Yi Liu and Peng Liu and Lin Ma and Le Weng and Xiaohang Hu and Xin Ma and Qian Qian and Rongfei Jia and Binqiang Zhao and Hao Helen Zhang},
  journal={2021 IEEE/CVF International Conference on Computer Vision (ICCV)},
  year={2020},
  pages={10913-10922},
  url={https://api.semanticscholar.org/CorpusID:227013144}
}

@article{khanna2023hssd,
  title={{Habitat Synthetic Scenes Dataset (HSSD-200): An Analysis of 3D Scene Scale and Realism Tradeoffs for ObjectGoal Navigation}},
  author={Mukul Khanna and Yongsen Mao and Hanxiao Jiang and Sanjay Haresh and Brennan Schacklett and Dhruv Batra and Alexander Clegg and Eric Undersander and Angel X. Chang and Manolis Savva},
  journal={2024 IEEE/CVF Conference on Computer Vision and Pattern Recognition (CVPR)},
  year={2023},
  pages={16384-16393},
  url={https://api.semanticscholar.org/CorpusID:259203445}
}

@article{yang_holodeck_2024,
  title={{Holodeck: Language Guided Generation of 3D Embodied AI Environments}},
  author={Yue Yang and Fan-Yun Sun and Luca Weihs and Eli VanderBilt and Alvaro Herrasti and Winson Han and Jiajun Wu and Nick Haber and Ranjay Krishna and Lingjie Liu and Chris Callison-Burch and Mark Yatskar and Aniruddha Kembhavi and Christopher Clark},
  journal={2024 IEEE/CVF Conference on Computer Vision and Pattern Recognition (CVPR)},
  year={2023},
  pages={16277-16287},
  url={https://api.semanticscholar.org/CorpusID:266210109}
}

@article{Para2020GenerativeLM,
  title={{Generative Layout Modeling using Constraint Graphs}},
  author={Wamiq Reyaz Para and Paul Guerrero and Tom Kelly and Leonidas J. Guibas and Peter Wonka},
  journal={2021 IEEE/CVF International Conference on Computer Vision (ICCV)},
  year={2020},
  pages={6670-6680},
  url={https://api.semanticscholar.org/CorpusID:227209310}
}

@inproceedings{cCelen2024IDesignPL,
  title={{I-Design: Personalized LLM Interior Designer}},
  author={{\c{C}}elen, Ata and Han, Guo and Schindler, Konrad and Van Gool, Luc and Armeni, Iro and Obukhov, Anton and Wang, Xi},
  booktitle={European Conference on Computer Vision},
  pages={217--234},
  year={2024},
  organization={Springer},
  url={https://api.semanticscholar.org/CorpusID:268876421}
}

@article{Shabani2022HouseDiffusionVF,
  title={{HouseDiffusion: Vector Floorplan Generation via a Diffusion Model with Discrete and Continuous Denoising}},
  author={Mohammad Amin Shabani and Sepidehsadat Hosseini and Yasutaka Furukawa},
  journal={2023 IEEE/CVF Conference on Computer Vision and Pattern Recognition (CVPR)},
  year={2022},
  pages={5466-5475},
  url={https://api.semanticscholar.org/CorpusID:254018175}
}

@inproceedings{Lin2024InstructScene,
title={{InstructScene: Instruction-Driven 3D Indoor Scene Synthesis with Semantic Graph Prior}},
author={Chenguo Lin and Yadong Mu},
booktitle={The Twelfth International Conference on Learning Representations},
year={2024},
url={https://openreview.net/forum?id=LtuRgL03pI}
}

@article{Fan2021FloorPlanCAD,
  title={{FloorPlanCAD: A Large-Scale CAD Drawing Dataset for Panoptic Symbol Spotting}},
  author={Zhiwen Fan and Lingjie Zhu and Honghua Li and Xiaohao Chen and Siyu Zhu and Ping Tan},
  journal={2021 IEEE/CVF International Conference on Computer Vision (ICCV)},
  year={2021},
  pages={10108-10117},
  url={https://api.semanticscholar.org/CorpusID:234742455}
}

@article{Deitke2022ProcTHOR,
  title={{ProcTHOR: Large-Scale Embodied AI Using Procedural Generation}},
  author={Deitke, Matt and VanderBilt, Eli and Herrasti, Alvaro and Weihs, Luca and Ehsani, Kiana and Salvador, Jordi and Han, Winson and Kolve, Eric and Kembhavi, Aniruddha and Mottaghi, Roozbeh},
  journal={Advances in Neural Information Processing Systems},
  volume={35},
  pages={5982--5994},
  year={2022},
  url={https://api.semanticscholar.org/CorpusID:249642405}
}

@article{Tang2024DiffuScene,
  title={{DiffuScene: Denoising Diffusion Models for Generative Indoor Scene Synthesis}},
  author={Jiapeng Tang and Yinyu Nie and Lev Markhasin and Angela Dai and Justus Thies and Matthias Nie{\ss}ner},
  journal={2024 IEEE/CVF Conference on Computer Vision and Pattern Recognition (CVPR)},
  year={2023},
  pages={20507-20518},
  url={https://api.semanticscholar.org/CorpusID:268363865}
}

@article{Le2022CodeRL,
  title={{CodeRL: Mastering Code Generation through Pretrained Models and Deep Reinforcement Learning}},
  author={Le, Hung and Wang, Yue and Gotmare, Akhilesh Deepak and Savarese, Silvio and Hoi, Steven Chu Hong},
  journal={Advances in Neural Information Processing Systems},
  volume={35},
  pages={21314--21328},
  year={2022},
  url={https://api.semanticscholar.org/CorpusID:250280117}
}

@article{Shao2024DeepSeekMath,
  title={{DeepSeekMath: Pushing the Limits of Mathematical Reasoning in Open Language Models}},
  author={Shao, Zhihong and Wang, Peiyi and Zhu, Qihao and Xu, Runxin and Song, Junxiao and Bi, Xiao and Zhang, Haowei and Zhang, Mingchuan and Li, YK and Wu, Yang and others},
  journal={ArXiv},
  year={2024},
  volume={abs/2402.03300},
  url={https://api.semanticscholar.org/CorpusID:267412607}
}

@inproceedings{Dou2024StepCoder,
  title={{StepCoder: Improving Code Generation with Reinforcement Learning from Compiler Feedback}},
  author={Dou, Shihan and Liu, Yan and Jia, Haoxiang and Zhou, Enyu and Xiong, Limao and Shan, Junjie and Huang, Caishuang and Wang, Xiao and Fan, Xiaoran and Xi, Zhiheng and others},
  booktitle={Proceedings of the 62nd Annual Meeting of the Association for Computational Linguistics (Volume 1: Long Papers)},
  pages={4571--4585},
  year={2024},
  url={https://api.semanticscholar.org/CorpusID:271915494}
}

@article{Fu20203DFUTURE,
  title={{3D-FUTURE: 3D Furniture Shape with TextURE}},
  author={Huan Fu and Rongfei Jia and Lin Gao and Mingming Gong and Binqiang Zhao and Stephen J. Maybank and Dacheng Tao},
  journal={International Journal of Computer Vision},
  year={2020},
  volume={129},
  pages={3313 - 3337},
  url={https://api.semanticscholar.org/CorpusID:221819358}
}

@article{Feng2023LayoutGPT,
  title={{LayoutGPT: Compositional Visual Planning and Generation with Large Language Models}},
  author={Feng, Weixi and Zhu, Wanrong and Fu, Tsu-jui and Jampani, Varun and Akula, Arjun and He, Xuehai and Basu, Sugato and Wang, Xin Eric and Wang, William Yang},
  journal={Advances in Neural Information Processing Systems},
  volume={36},
  pages={18225--18250},
  year={2023}
}

@article{Hu2020Graph2Plan,
  title={{Graph2Plan: Learning Floorplan Generation from Layout Graphs}},
  author={Ruizhen Hu and Zeyu Huang and Yuhan Tang and Oliver Matias van Kaick and Hao Zhang and Hui Huang},
  journal={ACM Transactions on Graphics (TOG)},
  year={2020},
  volume={39},
  pages={118:1 - 118:14},
  url={https://api.semanticscholar.org/CorpusID:216562245}
}

@article{Leimer2022LayoutEnhancer,
  title={{LayoutEnhancer: Generating Good Indoor Layouts from Imperfect Data}},
  author={Kurt Leimer and Paul Guerrero and Tomer Weiss and Przemyslaw Musialski},
  journal={SIGGRAPH Asia 2022 Conference Papers},
  year={2022},
  url={https://api.semanticscholar.org/CorpusID:252734701}
}

@article{Merrell2011InteractiveFurniture,
  title={{Interactive furniture layout using interior design guidelines}},
  author={Paul C. Merrell and Eric Schkufza and Zeyang Li and Maneesh Agrawala and Vladlen Koltun},
  journal={ACM SIGGRAPH 2011 papers},
  year={2011},
  url={https://api.semanticscholar.org/CorpusID:53246134}
}

@inproceedings{Nauata2020HouseGAN,
  title={{House-GAN: Relational Generative Adversarial Networks for Graph-constrained House Layout Generation}},
  author={Nauata, Nelson and Chang, Kai-Hung and Cheng, Chin-Yi and Mori, Greg and Furukawa, Yasutaka},
  booktitle={European Conference on Computer Vision},
  pages={162--177},
  year={2020},
  organization={Springer},
  url={https://api.semanticscholar.org/CorpusID:212725507}
}

@inproceedings{Paschalidou2021ATISS,
  title={{ATISS: Autoregressive Transformers for Indoor Scene Synthesis}},
  author={Despoina Paschalidou and Amlan Kar and Maria Shugrina and Karsten Kreis and Andreas Geiger and Sanja Fidler},
  booktitle={Neural Information Processing Systems},
  year={2021},
  url={https://api.semanticscholar.org/CorpusID:238419213}
}

@article{Wu2019RPLAN,
  title={{Data-driven interior plan generation for residential buildings}},
  author={Wenming Wu and Xiaoming Fu and Rui Tang and Yuhan Wang and Yuanhang Qi and Ligang Liu},
  journal={ACM Transactions on Graphics (TOG)},
  year={2019},
  volume={38},
  pages={1 - 12},
  url={https://api.semanticscholar.org/CorpusID:207998029}
}

@article{Yang2025OptiScene,
  title={{LLM-driven Indoor Scene Layout Generation via Scaled Human-aligned Data Synthesis and Multi-Stage Preference Optimization}},
  author={Yixuan Yang and Zhen Luo and Tongsheng Ding and Junru Lu and Mingqi Gao and Jinyu Yang and Victor Sanchez and Feng Zheng},
  journal={Advances in Neural Information Processing Systems},
  volume={38},
  pages={42499--42529},
  year={2025},
  url={https://api.semanticscholar.org/CorpusID:279251590}
}

@article{Yu2011MakeItHome,
  title={{Make it home: Automatic Optimization of Furniture Arrangement.}},
  author={Yu, Lap-Fai and Yeung, Sai Kit and Tang, Chi-Keung and Terzopoulos, Demetri and Chan, Tony F and Osher, Stanley J},
  journal={ACM SIGGRAPH 2011 papers},
  volume={30},
  number={4},
  pages={86},
  year={2011},
  url={https://api.semanticscholar.org/CorpusID:14227}
}

@inproceedings{Zheng2020Structured3D,
  title={{Structured3D: A Large Photo-realistic Dataset for Structured 3D Modeling}},
  author={Zheng, Jia and Zhang, Junfei and Li, Jing and Tang, Rui and Gao, Shenghua and Zhou, Zihan},
  booktitle={European Conference on Computer Vision},
  pages={519--535},
  year={2020},
  organization={Springer},
  url={https://api.semanticscholar.org/CorpusID:199064623}
}

@article{sun2024layoutvlm,
  title={{LayoutVLM: Differentiable Optimization of 3D Layout via Vision-Language Models}},
  author={Fan-Yun Sun and Weiyu Liu and Siyi Gu and Dylan Lim and Goutam Bhat and Federico Tombari and Manling Li and Nick Haber and Jiajun Wu},
  journal={2025 IEEE/CVF Conference on Computer Vision and Pattern Recognition (CVPR)},
  year={2024},
  pages={29469-29478},
  url={https://api.semanticscholar.org/CorpusID:274446060}
}

@inproceedings{Pfaff2026SceneSmithAG,
  title={{SceneSmith: Agentic Generation of Simulation-Ready Indoor Scenes}},
  author={Pfaff, Nicholas and Cohn, Thomas and Zakharov, Sergey and Cory, Rick and Tedrake, Russ},
  booktitle={Forty-third International Conference on Machine Learning},
  year={2026},
  url={https://api.semanticscholar.org/CorpusID:285463147}
}

@article{Liu2025UnderstandingRT,
  title={{Understanding R1-Zero-Like Training: A Critical Perspective}},
  author={Zichen Liu and Changyu Chen and Wenjun Li and Penghui Qi and Tianyu Pang and Chao Du and Wee Sun Lee and Min Lin},
  journal={ArXiv},
  year={2025},
  volume={abs/2503.20783},
  url={https://api.semanticscholar.org/CorpusID:277322777}
}

@article{DeepSeekAI2025DeepSeekR1IR,
  title={{DeepSeek-R1: Incentivizing Reasoning Capability in LLMs via Reinforcement Learning}},
  author={DeepSeek-AI},
  journal={ArXiv},
  year={2025},
  volume={abs/2501.12948},
  url={https://api.semanticscholar.org/CorpusID:284488789}
}

@article{Tam2025SceneEvalES,
  title={{SceneEval: Evaluating Semantic Coherence in Text-Conditioned 3D Indoor Scene Synthesis}},
  author={Hou In Ivan Tam and Hou In Derek Pun and Austin T. Wang and Angel X. Chang and Manolis Savva},
  journal={2026 IEEE/CVF Winter Conference on Applications of Computer Vision (WACV)},
  year={2025},
  pages={7355-7365},
  url={https://api.semanticscholar.org/CorpusID:277113092}
}

@inproceedings{Delalandre2011RecentCO,
  title={{Recent contributions on the SESYD dataset for performance evaluation of symbol spotting systems}},
  author={Mathieu Delalandre and Ernest Valveny and Jean-Yves Ramel},
  year={2011},
  url={https://api.semanticscholar.org/CorpusID:7882321}
}

@misc{gemmateam2026gemma4,
      title={{Gemma 4 Technical Report}}, 
      author={{Gemma Team}},
      year={2026},
      eprint={2607.02770},
      archivePrefix={arXiv},
      primaryClass={cs.CL},
      url={https://arxiv.org/abs/2607.02770}, 
}

@article{Hu2021LoRALA,
  title={{LoRA: Low-Rank Adaptation of Large Language Models}},
  author={Edward J. Hu and Yelong Shen and Phillip Wallis and Zeyuan Allen-Zhu and Yuanzhi Li and Shean Wang and Weizhu Chen},
  journal={ArXiv},
  year={2021},
  volume={abs/2106.09685},
  url={https://api.semanticscholar.org/CorpusID:235458009}
}

@techreport{anthropic2026sonnet5systemcard,
  author      = {{Anthropic}},
  title       = {{Claude Sonnet 5 System Card}},
  institution = {Anthropic},
  year        = {2026},
  month       = {June},
  url         = {https://www.anthropic.com/claude-sonnet-5-system-card},
}

@misc{moonshot2026kimik3,
  author = {{Moonshot AI}},
  title  = {{Kimi K3: Open Frontier Intelligence}},
  year   = {2026},
  month  = {July},
  url    = {https://www.kimi.ai/blog/kimi-k3}
}

@inproceedings{Loshchilov2017DecoupledWD,
  title={{Decoupled Weight Decay Regularization}},
  author={Ilya Loshchilov and Frank Hutter},
  booktitle={International Conference on Learning Representations},
  year={2017},
  url={https://api.semanticscholar.org/CorpusID:53592270}
}

@misc{blender,
  author = {{Blender Online Community}},
  title  = {{Blender - a 3D modelling and rendering package}},
  organization = {Blender Foundation, Blender Institute},
  address = {Amsterdam},
  url    = {http://www.blender.org}
}

@misc{openai2025gpt52,
  author = {{OpenAI}},
  title  = {Introducing {GPT}-5.2},
  year   = {2025},
  month  = {December},
  url    = {https://openai.com/index/introducing-gpt-5-2/}
}
\bibliographystyle{plainnat}

\clearpage
\appendix
\section{AntPlan dataset}
\label{app:dataset}

AntPlan pairs original architectural floor-plan images with structural, room, and object annotations. Its vocabulary comprises ten room categories and 92 object classes, covering furniture, fixtures, appliances, accessories, and service elements. The original drawings and separate annotation layers support both architectural drawing understanding and the construction of room-level furnishing examples.

\begin{figure}[htbp]
  \centering
  \includegraphics[width=\linewidth]{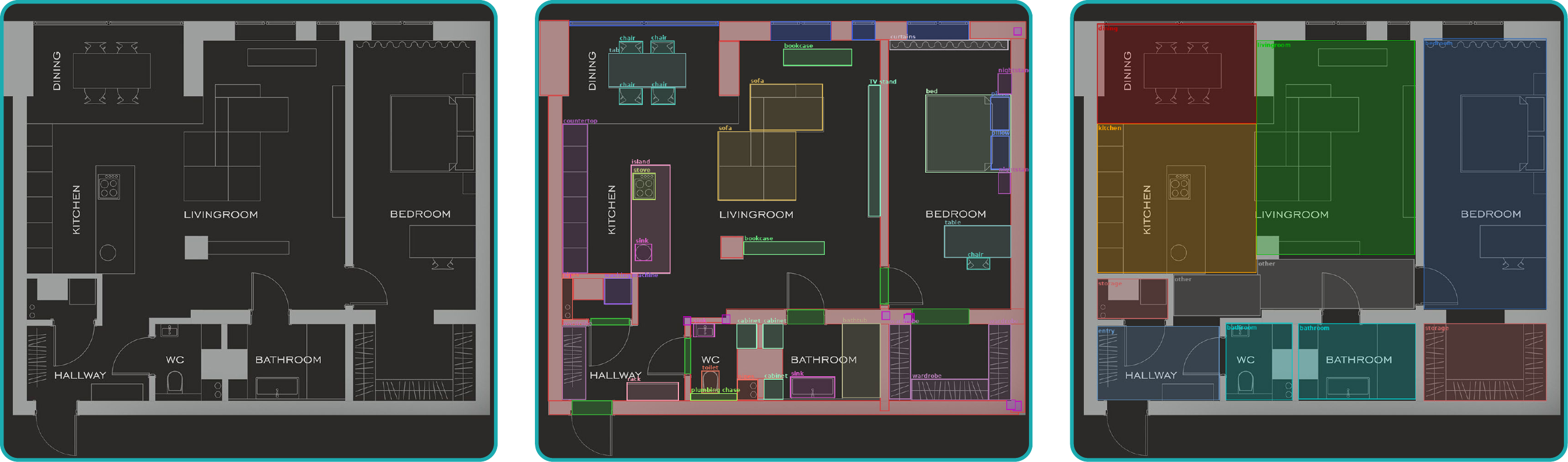}
  \caption{An annotated AntPlan example. Left: the original architectural drawing. Center: structural and object bounding boxes with semantic labels. Right: annotated room regions and room categories.}
  \label{fig:antplan-211}
\end{figure}

\subsection{Annotation workflow}
\label{app:dataset-labeling}

The plans are collected from publicly accessible sources and retain their original drawing styles. Structural annotation begins with an RT-DETR-X detector~\citep{Lv2024RTDETRv2IB} trained on CubiCasa5K~\citep{kalervo2019cubicasa5k}, extracting walls, doors, windows, and railings. Room regions and their semantic categories are labeled manually. A separate furniture detector is bootstrapped from a hand-labeled subset and applied to the remaining drawings, followed by manual review and correction. This workflow combines detector proposals with human annotation rather than requiring every object box to be drawn from scratch.

The annotation layers describe different aspects of the same image. Room labels identify functional regions, including separate kitchen, dining, and living regions within an open-plan interior. Object labels specify a category and an axis-aligned bounding box for each annotated instance. Structural boxes describe the building geometry and openings independently of movable furniture. The saved structural layer also carries linkage-point annotations; these are retained in the data records but are not included in the furnishing model's text input.

Figure~\ref{fig:antplan-211} illustrates these layers on one floor plan. The object vocabulary distinguishes furniture and appliances from details such as radiators, pipes, and plumbing chases that can affect usable space. Detector-assisted annotations remain subject to missed symbols, ambiguous classes, and imperfect box boundaries. We therefore treat the reviewed furniture annotations as pseudo-labels rather than claiming exhaustive annotation accuracy.

\subsection{Dataset statistics}
\label{app:dataset-statistics}

The cleaned annotation snapshot contains 505 plans with room labels, totaling 4,495 room annotations and 26,716 object annotations. This corresponds to 8.90 rooms per room-labeled plan and 53.01 objects per object-labeled plan. Of the object annotations, 26,001 are associated with a room and 715 remain unassigned. Table~\ref{tab:antplan-room-statistics} reports the room-associated counts before the filtering used for model adaptation.

\begin{table}[htbp]
\centering\small
\setlength{\belowcaptionskip}{5pt}
\caption{AntPlan annotations by room category, before experimental filtering. Object counts include all annotated classes assigned to each room, including accessories and service elements. Observed class counts overlap across categories and are not the model whitelists. Unassigned objects are excluded from this table.}
\label{tab:antplan-room-statistics}
\begin{tabular}{lrrrr}
\toprule
Room category & Rooms & Objects & Objects/room & Observed classes \\
\midrule
Bedroom & 790 & 7,048 & 8.92 & 67 \\
Bathroom & 769 & 4,145 & 5.39 & 51 \\
Kitchen & 486 & 5,485 & 11.29 & 56 \\
Living room & 550 & 5,097 & 9.27 & 61 \\
Dining room & 80 & 720 & 9.00 & 38 \\
Entry & 457 & 937 & 2.05 & 43 \\
Storage & 443 & 1,226 & 2.77 & 48 \\
Balcony & 264 & 745 & 2.82 & 48 \\
Terrace & 33 & 38 & 1.15 & 10 \\
Other & 623 & 560 & 0.90 & 60 \\
\midrule
Total & 4,495 & 26,001 & 5.78 & 92 \\
\bottomrule
\end{tabular}
\end{table}


\subsection{Saved records and room extraction}
\label{app:dataset-records}

Each floor plan is linked by a shared identifier to its original image and separate object, room, and structural label files. The label files store a class index and normalized bounding-box center, width, and height; the accompanying class dictionaries resolve indices to semantic names. A combined JSON record stores the image dimensions, structural boxes, room categories and boxes, objects associated with each room, and any unassigned objects. At this full-plan level, boxes use image-pixel coordinates in the order $(x_{\min},y_{\min},x_{\max},y_{\max})$. Saved overlays and schematic views provide visual checks of these records.

For the floor plan in Figure~\ref{fig:antplan-211}, the saved record links the original drawing to ten room regions and 38 object annotations. One of its room entries has class \texttt{bedroom}, with a \texttt{bed} object whose pixel box is $[1309.2,206.2,1496.2,376.2]$. These full-plan records preserve the original annotations; the furnishing examples are derived from them.

For each eligible room, we crop its annotated region with a 0.5\,m context margin, limited by the image boundary. Pixel geometry is converted to room-local metric coordinates using a scale estimated from the median annotated door-opening width, assuming a width of 0.8\,m. This is an approximate scale calibration rather than a measurement recovered from the drawing's dimension text. Structural and object boxes are clipped to the crop, translated to its top-left origin, and rounded to two decimal places in meters. Rooms from plans without a usable door-based scale cannot enter this extraction stage.

A derived room record retains its parent identifier and room index, room category, scale estimate, crop bounds, room bounds, structural geometry, and furniture boxes. For instance, the kitchen in Figure~\ref{fig:antplan-211} has an extracted frame measuring $4.37\times4.14$\,m and contains five object annotations. The training record links this geometry to a structure-only image, a furniture request, and the target DSL. Object classes, counts, and approximate sizes form the request, while their positions form the supervision. The structure-only model image omits the target furniture, although the original furnished image remains part of the dataset.

\subsection{Experimental splits and class whitelists}
\label{app:dataset-splits}

The furnishing experiments use bedroom, bathroom, kitchen, and living-room subsets of the cleaned annotations. For each modeled room category, object labels are normalized to DSL class names and filtered by the corresponding whitelist in Table~\ref{tab:antplan-whitelists}. A bathroom must retain at least one object; the other three room types must retain at least two. The four whitelists contain 53 distinct DSL classes in total. These filters define the experimental subsets and do not remove classes from the full annotation vocabulary.

Within each room type, all examples from one source plan belong to the same training or validation split. The partitions are room-specific: a source plan can occur in training for one adapter and validation for another. The direct-DSL and reasoning prompt variants retain the same room membership, structure-only images, furniture requests, and target layouts. The original annotation-layer train/validation folders and these room-level furnishing splits serve different tasks and should not be conflated.

\begin{table}[H]
\centering\small
\setlength{\belowcaptionskip}{5pt}
\caption{Room and target-object counts in the furnishing splits after class filtering. ``Excluded'' counts rooms below the minimum retained-object threshold, after source-plan and room extraction filters. Whitelist sizes describe permitted DSL classes, including classes absent from a particular split; the total counts distinct classes across room types.}
\label{tab:dataset-subsets}
\setlength{\tabcolsep}{5pt}
\begin{tabular}{lrrrrrr}
\toprule
 & \multicolumn{2}{c}{Rooms} & \multicolumn{2}{c}{Target objects} & & \\
\cmidrule(lr){2-3}\cmidrule(lr){4-5}
Room type & Train & Val. & Train & Val. & Excluded & Whitelist \\
\midrule
Bedroom & 672 & 66 & 4,842 & 485 & 19 & 26 \\
Bathroom & 677 & 74 & 3,586 & 369 & 1 & 17 \\
Kitchen & 428 & 47 & 4,677 & 534 & 1 & 26 \\
Living room & 482 & 56 & 3,945 & 464 & 0 & 21 \\
\midrule
Total & 2,259 & 243 & 17,050 & 1,852 & 21 & 53 \\
\bottomrule
\end{tabular}
\end{table}

\begin{table}[H]
\centering\small
\setlength{\belowcaptionskip}{5pt}
\caption{Exact room-specific object whitelists used for the furnishing experiments. Spaces are shown for readability; DSL class names use underscores and lowercase letters. The kitchen policy permits both \texttt{island} and \texttt{kitchen\_island}. These are permitted model vocabularies, distinct from the full 92-class annotation vocabulary.}
\label{tab:antplan-whitelists}
\renewcommand{\arraystretch}{1.2}
\begin{tabularx}{\linewidth}{@{}l>{\raggedright\arraybackslash}X@{}}
\toprule
Room type & Permitted object classes \\
\midrule
Bedroom & armchair, bed, bench, bookcase, chair, closet, coffee table, curtains, dressing table, floor lamp, lamp, laptop, monitor, nightstand, plant, pouf, rack, radiator, rug, shelf, sofa, table, tv, tv stand, wardrobe, workstation \\
\addlinespace
Bathroom & bathtub, bidet, boiler, cabinet, closet, dryer, pipes, plumbing chase, radiator, service void, shelf, shower, sink, toilet, toilet paper, towel warmer, washing machine \\
\addlinespace
Kitchen & bar counter, cabinet, chair, closet, countertop, curtains, dishwashing machine, fridge, island, kitchen island, microwave, oven, pipes, rack, radiator, range hood, rug, service void, shelf, sink, stove, table, tv, wardrobe, washing machine, wine cabinet \\
\addlinespace
Living room & armchair, bookcase, cabinet, chair, chimney, closet, coffee table, curtains, floor lamp, plant, pouf, rack, radiator, rug, shelf, side table, sofa, table, tv, tv stand, wardrobe \\
\bottomrule
\end{tabularx}
\end{table}


\section{Training and reward implementation}
\label{app:training-details}

\paragraph{Data and supervised training.}
The cleaned training/validation sets contain 672/66 bedrooms, 428/47 kitchens,
677/74 bathrooms, and 482/56 living rooms (2,259/243 room examples in total).
Each room type has its own split, disjoint at the parent-floorplan level
within that type; source plans can cross partitions between separately trained
room types. These training runs use no geometric augmentation. Each example includes a structure-only raster, its textual geometric
description, the requested inventory, and the annotated furniture DSL.
We train a separate language-model LoRA adapter for each room type; the
vision encoder is frozen. SFT minimizes next-token cross-entropy over the
DSL and turn-end token, normalized by the number of supervised tokens across
each accumulation window. Prompt, image, and padding tokens are masked.

\begin{table}[ht]
\centering\small
\setlength{\belowcaptionskip}{5pt}
\caption{Training settings for layout SFT followed by GRPO with reasoning.}
\label{tab:training_recipe}
\begin{tabular}{p{0.29\linewidth}p{0.63\linewidth}}
\toprule
Setting & Value \\
\midrule
Backbone & Gemma-4-31B-it; BF16 \\
LoRA & Rank 64, scaling 128, dropout 0.05; language attention and MLP projections \\
SFT & 5 epochs; batch 1, accumulation 8; AdamW~\citep{Loshchilov2017DecoupledWD}, weight decay 0 \\
SFT learning rate & $2\times10^{-5}$; 5\% warmup, cosine decay to $2\times10^{-6}$ \\
GRPO with reasoning & 15 updates; 8 rooms per update, 16 completions per room \\
GRPO optimizer & AdamW; learning rate $2\times10^{-5}$, cosine decay to zero \\
GRPO regularization & Frozen-base KL, $\beta=0.001$; advantage clipping at $\pm2$ \\
Gradient clipping & Global norm 1.0 in both stages \\
Decoding & Temperature 1.0, top-$p$ 0.95; maximum new tokens: kitchen 6,144; living room 4,096; bathroom and bedroom 3,072 \\
Validation & 8 samples per held-out room; seed 1234 \\
\bottomrule
\end{tabular}
\end{table}

\paragraph{Training reward.}
The final furniture block is extracted after removing completed reasoning blocks. The reasoning text receives no separate quality or format reward. The reported LRS $S$ starts at 10 and adds the applicable negative rule
deltas. It is distinct from the optimization reward. Let $B$ count gross size
mismatches, $q=S-2B$, and $G$ denote reference-layout similarity. We use
\begin{equation}
 R(y)=\begin{cases}
 -40,&\text{if no furniture objects can be parsed},\\
 q+2\,\sigma((q-7)/1.5)G,&\text{otherwise}.
 \end{cases}
\label{eq:training_reward_detail}
\end{equation}
Here $\sigma$ is the logistic sigmoid. This discourages empty answers and objects shrunk to evade geometric checks.
An unfinished answer is penalized through the same parsing criterion; truncation
alone does not override an otherwise parseable layout.
For a predicted and reference object, sort their dimensions into short and long
sides $(s,l)$ and $(s',l')$, and define
\begin{equation}
 f=\sqrt{\frac{\min(s,s')}{\max(s,s')}\frac{\min(l,l')}{\max(l,l')}}.
\end{equation}
For $B$, objects are paired within each class by increasing area, and a pair
contributes one when $f<0.5$; missing and surplus objects are handled by the
inventory penalty. For $G$, reference objects are greedily paired to the nearest
unused predicted center of the same class:
\begin{equation}
 G=\frac{1}{n_{\rm ref}}\sum_{j=1}^{n_{\rm ref}}
 \exp\!\left(-\frac{\|c_j-\hat c_{m(j)}\|_2}{0.8}\right)f_{j,m(j)},
\end{equation}
with unmatched reference objects contributing zero. Distances are in meters.
The sigmoid therefore gates agreement with one annotated arrangement by basic
feasibility; it does not make that arrangement the unique valid solution.

\paragraph{Dr.~GRPO-style group-relative update.}
At each iteration we sample fresh reasoning-plus-layout completions from the current policy and make
one optimizer update. For group $g$ of 16 answers, advantages are
\begin{equation}
 A_{gi}=\operatorname{clip}\!\left(
 \frac{R_{gi}-\overline R_g}{\sigma_{\rm batch}},-2,2\right),
\end{equation}
where $\sigma_{\rm batch}$ is the population standard deviation across the
current iteration's 128 rewards. Groups with identical rewards are skipped.
This is a group-relative policy-gradient implementation; it does not use a
PPO-clipped importance-ratio surrogate. To specify the implemented accumulation
exactly, let $C$ be a token-budget microbatch, $N$ the number of retained answers,
$m_{it}$ the completion-token mask,
$o_i=(r_i,y_i)$ the complete reasoning-plus-layout response, and $\ell_{it}=\log\pi_\theta(o_{it}\mid x_i,o_{i,<t})$. The accumulated loss is
\begin{align}
 L&=\sum_C\frac{|C|}{N}\left[
 -\frac{1}{1024}\sum_{i\in C,t}A_i m_{it}\ell_{it}
 +\beta\frac{\sum_{i\in C,t}m_{it}(e^{d_{it}}-d_{it}-1)}
 {\sum_{i\in C,t}m_{it}}\right],\\
 d_{it}&=\operatorname{clip}(\ell^{\rm base}_{it}-\ell_{it},-20,20).
\end{align}

\section{Layout Rule Score}
\label{app:scoring_details}
The score aggregates explicit geometric and functional constraints inferred from the ground-truth annotations of professional floor plans. Together, they provide a proxy for plausible furniture arrangements. We refer to this aggregate as the Layout Rule Score (LRS), denoted by $S$. Its use of annotated plans to inform the rules is distinct from the per-example ground-truth similarity term added to the training reward: evaluating $S$ requires only the proposed layout and room input, not a reference furniture arrangement. The following checks, tolerances, and penalties specify the implemented proxy.


Rules operate in meters on semantic footprints and structural geometry. They
use class-dependent exceptions for legitimate plan-view overlap, including
rugs under furniture, a television on its stand, and built-in appliances.
Table~\ref{tab:rule_families} summarizes the shared checks; room-specific checks
follow below. Small annotation tolerances are intended to suppress extraction
noise, and should not be read as building-code requirements.

\begin{table}[ht]
\centering\small
\setlength{\belowcaptionskip}{5pt}
\caption{Principal LRS penalty families. Costs are deducted from 10; multiple violations can accumulate.}
\label{tab:rule_families}
\begin{tabular}{p{0.22\linewidth}p{0.69\linewidth}}
\toprule
Check & Criterion and cost \\
\midrule
Empty layout & No parsed objects: 15, giving reported score $-5$ \\
Inventory & 2 per missing or surplus requested item, capped at 8 \\
Bounds and room & 2 per out-of-frame object; 2 when more than half lies beyond a railing-defined room boundary \\
Wall collision & Object-area overlap: cost 1 for light contact of at least $0.04\,\mathrm{m}^2$, 3 at 10--30\%, 4 above 30\%; penetrating wall: an additional 3 \\
Opening obstruction & Door overlap 2; window overlap 1; 0.60\,m door-clearance obstruction 2. Overlap tolerances are 5\% for openings and 10\% for clearance zones \\
Furniture collision & For intersection $a$, let $r_s=a/A_{\min}$ and $r_l=a/A_{\max}$. Ignore $a<0.04\,\mathrm{m}^2$ with $r_s<0.25$, or $r_s<0.10$. Otherwise cost 4 if $r_l\ge0.50$, 2 if $r_s\ge0.15$, and 1 otherwise; designated forbidden pairs incur severe cost \\
Wall placement & Required fixtures within 0.15\,m of a wall; default failure cost 1, with room-specific variants below \\
Accessibility & One point per additional usable connected region and one per vertex in a minimum vertex cover of the blocked-target graph \\
\bottomrule
\end{tabular}
\end{table}

\paragraph{Accessibility.}
We subtract walls, railings, and structural voids from the room frame, use the
largest remaining floor polygon, then erode the floor and inflate furniture
obstacles by clearance $c$. Here $c=0.20$\,m for bedrooms, living rooms and
kitchens, and $c=0.10$\,m for bathrooms, corresponding to nominal passage
widths of 0.40 and 0.20\,m. Remaining components smaller than
$0.10\,\mathrm{m}^2$ are discarded. Furniture targets and valid door openings
are reachable when they access a common component; targets separated by at
most 0.05\,m are treated as adjacent. Targets thinner than 0.25\,m are filtered,
and appliances at least 90\% embedded in a host surface are not separate
destinations. Doors are checked with inward reach beginning at 0.60\,m and
extending to 0.90/1.20\,m when needed. Movable chairs and wall-hung or
walked-over objects are excluded from obstacles. The resulting blocked-pair
graph is summarized by a minimum vertex cover to avoid repeatedly charging
for one isolated object. These are conservative geometric proxies for access,
not full human-motion or usability simulation.

\paragraph{Room-specific checks.}
Kitchens require a 0.70\,m clear aisle on at least one island side (cost 1 if
none exists), a 0.60\,m table--counter gap (cost ten times a positive gap
deficit), and 0.60\,m work/swing zones at appliances (cost 1 above 20\%
blockage). Same-side chairs need 0.05\,m separation; intermediate fitted-run
gaps of 0.05--0.30\,m cost 1. Wall anchoring may be satisfied through a fitted
host for a sink or fridge. Bathrooms apply cost 3 for required fixtures more
than 0.15\,m from a wall, with hosting/adjacency exceptions; partially mounted
sink--cabinet and washing-machine--cabinet pairs cost 2 when containment is
between 5\% and 80\%. In bedrooms, bed/wardrobe/closet wall embedding of at
least 10\% receives the severe wall cost; the revised off-wall cost for these
classes grows linearly from 1 at 0.15\,m to 4 at 0.50\,m and remains 4 beyond.
Other wall-required bedroom classes retain the flat cost. Living rooms check
that the television intersects the sofa's facing corridor (cost 1 otherwise)
and reject excessive overlap between components of a sofa. Storage anchoring,
window-front obstruction, and legitimate mounting/overlap exceptions are
defined by room-specific semantic class sets.


\section{Training variants and outcome-supervised reasoning}
\label{app:reasoning}

\paragraph{Choice of configurations.}
Table~\ref{tab:training_ablation} follows the pipeline from the pretrained model through supervised adaptation to reinforcement learning. Both the base and SFT stages generate DSL directly, keeping the output format fixed for that comparison. The final stage adds GRPO with reasoning and room-specific completion allowances, so its gain does not isolate the effect of reasoning alone. Table~\ref{tab:base-input-ablation} retains the stronger reasoning-prompted base reference, and Table~\ref{tab:training-input-ablation} includes direct-DSL GRPO to compare reinforcement learning without a reasoning prefix.

\paragraph{Learning reasoning from the final layout.}
The second stage applies GRPO to responses containing a free-form \texttt{<think>} block followed by the final \texttt{FURNITURE} block. Following the outcome-supervision principle of DeepSeek-R1-Zero, no target trace or step-level reasoning reward is supplied. The model starts from our layout-SFT adapter, generates its own intermediate reasoning, and receives a scalar reward computed from the final furniture arrangement. That reward trains the full response, including the reasoning tokens. LRS provides verifiable geometric and functional feedback, although it remains a proxy for overall design quality rather than a complete correctness test.

The prompt asks the model to consider alternative arrangements but prescribes no internal reasoning structure. It requires a separation between reasoning and the final DSL answer so that the latter can be parsed. We do not train a reasoning-quality evaluator or separately reward the reasoning format. A high-scoring answer can reinforce the trajectory that produced it without establishing that every statement in the trace is correct. The R1-Zero connection concerns outcome-only supervision, not an experimentally isolated benefit of intermediate reasoning.

\paragraph{Training comparison.}
Table~\ref{tab:training-input-ablation} combines SFT with direct DSL output, SFT followed by direct-DSL or reasoning GRPO, and reasoning GRPO initialized from the pretrained model without SFT. All conditions in this table use text and a structure-only image; DSL-only supervision specifies the output format, not the input modality. Table~\ref{tab:base-input-ablation} separately compares input and prompting conditions for the untuned base model.

Each condition is evaluated on the same held-out room IDs, with eight samples per room at temperature 1.0. We retain outputs with at least one parsed object and no reported parse errors. The mean score first averages retained samples within each room, then averages rooms with at least one retained sample; best-of-eight averages the highest retained score per room. The exclusion rate counts rejected samples over all generated samples. Kitchen SFT + GRPO with reasoning retains candidates in 43 of 47 rooms; the other measured conditions cover every validation room. Scores are taken from the saved LRS evaluations, without training-only reward terms.

GRPO without SFT uses 25 updates for kitchens, bathrooms, living rooms, and bedrooms; the SFT-initialized branches use 15 GRPO updates. These are comparisons of available training recipes, not equal-compute estimates of the isolated effect of SFT or reasoning.

SFT + GRPO with reasoning has higher retained-layout mean scores than direct-DSL GRPO in kitchens, bathrooms, and bedrooms, while direct-DSL GRPO performs better in living rooms. Direct-DSL GRPO also has higher best-of-eight scores in kitchens (5.91 versus 5.80) and living rooms (9.07 versus 8.95). Reasoning GRPO after SFT exceeds GRPO without SFT in both score summaries for every room type. These conditional scores must be read alongside generation reliability, particularly the 54.0\% excluded-sample rate for kitchen SFT + GRPO with reasoning.

\paragraph{Choice of the final model.}
The reasoning model offers a useful tradeoff for kitchens, where appliance placement, countertop relationships, and working clearances interact. On the 43 held-out kitchens with retained outputs from both branches, its mean LRS is 3.48 versus 1.66 for direct-DSL GRPO. The room-averaged fraction of retained layouts with disallowed overlaps falls from 36.6\% to 17.1\%, and with insufficient clearance from 18.9\% to 5.8\%; both branches place approximately 11.5 objects per room. These LRS component diagnostics support improved handling of interacting constraints among completed layouts, alongside the bathroom and bedroom gains in Table~\ref{tab:training-input-ablation}. They remain conditional on successful generation: wall-overlap violations increase from 30.2\% to 37.0\%, and the exclusion rates in the table remain a limitation.

High LRS does not certify complete layout validity. Spacious, sparsely furnished rooms can avoid collisions while still having poor functional relationships; a scalar score can also trade one violation against another. Inspection of training outputs revealed such tradeoffs, including wall-adjacent furniture moving away from walls while collision costs decreased. We therefore consider rule breakdowns and visual coherence alongside aggregate LRS; these observations do not establish that reasoning eliminates reward exploitation.

\paragraph{Room-specific generation budgets.}
Direct-DSL evaluations and GRPO without SFT use a 3,072-token completion allowance. For the final reasoning model after SFT, kitchens use 6,144 tokens, living rooms 4,096, and bathrooms and bedrooms 3,072. The larger allowances leave more room for reasoning and the final furniture block in the more complex kitchen and living-room layouts. We report these settings as part of each recipe rather than as a token-budget sweep.

\begin{table}[t]
\setlength{\belowcaptionskip}{5pt}
\caption{Training comparison on held-out AntPlan rooms, using eight samples per room with text and image input. Mean and best-of-eight LRS average over rooms after excluding empty outputs and outputs with parse errors; Excl. reports the percentage of all samples excluded. Kitchen SFT + GRPO with reasoning covers 43 of 47 rooms; every other measured condition covers all rooms. Reasoning outputs include a final DSL layout. Room-specific generation budgets and training settings are described in the text.}
\label{tab:training-input-ablation}
\label{tab:reasoning-ablation}
\centering\small
\setlength{\tabcolsep}{4pt}
\renewcommand{\arraystretch}{1.08}
\begin{tabular*}{\linewidth}{@{\extracolsep{\fill}}lllrrr@{}}
\toprule
Training & Output & Input & \shortstack{Mean\\LRS} & \shortstack{Best-of-8\\LRS} & Excl. (\%) \\
\midrule
\multicolumn{6}{l}{\textbf{Kitchen} (47 rooms)} \\
SFT & DSL & Text + image & $-0.75$ & $5.73$ & 0.3 \\
SFT + GRPO & DSL & Text + image & $1.27$ & $5.91$ & 0.3 \\
GRPO only & Reasoning & Text + image & $0.26$ & $5.27$ & 1.3 \\
SFT + GRPO & Reasoning & Text + image & $3.48$ & $5.80$ & 54.0 \\
\midrule
\multicolumn{6}{l}{\textbf{Bathroom} (74 rooms)} \\
SFT & DSL & Text + image & $6.25$ & $8.50$ & 0.2 \\
SFT + GRPO & DSL & Text + image & $6.87$ & $8.72$ & 0.2 \\
GRPO only & Reasoning & Text + image & $5.30$ & $8.46$ & 1.2 \\
SFT + GRPO & Reasoning & Text + image & $7.00$ & $9.00$ & 7.1 \\
\midrule
\multicolumn{6}{l}{\textbf{Living room} (56 rooms)} \\
SFT & DSL & Text + image & $5.41$ & $8.77$ & 0.0 \\
SFT + GRPO & DSL & Text + image & $6.79$ & $9.07$ & 0.0 \\
GRPO only & Reasoning & Text + image & $4.66$ & $8.43$ & 3.1 \\
SFT + GRPO & Reasoning & Text + image & $6.22$ & $8.95$ & 0.2 \\
\midrule
\multicolumn{6}{l}{\textbf{Bedroom} (66 rooms)} \\
SFT & DSL & Text + image & $5.71$ & $8.48$ & 0.0 \\
SFT + GRPO & DSL & Text + image & $5.32$ & $7.77$ & 0.0 \\
GRPO only & Reasoning & Text + image & $2.93$ & $7.58$ & 1.1 \\
SFT + GRPO & Reasoning & Text + image & $6.02$ & $8.79$ & 2.7 \\
\bottomrule
\end{tabular*}
\end{table}

\paragraph{Earlier attempts at trace supervision.}
The annotated professional plans provide furniture layouts but no reasoning traces. Earlier experiments explored procedural explanations, model-generated explanations conditioned on fixed layouts, and filtering rewarded rollouts for subsequent SFT. Constructing a reliable trace-supervision set proved difficult: plausible explanations could contain unsupported claims, and final-layout reward did not establish trace correctness. These experiments are separate from the current second stage, which learns reasoning directly through GRPO without constructing a trace-SFT dataset.

In one earlier bedroom experiment, procedural reasoning included final coordinates and was visible in the prompt while masked from the SFT loss; only the final DSL was supervised. Low teacher-forced loss could therefore reflect copying rather than useful reasoning, and free-generation quality declined. This finding concerns that trace-supervision setup. It does not establish a limitation of outcome-supervised GRPO, and it does not explain away the output-budget bottleneck in the later reasoning runs.

\paragraph{Limits of outcome feedback.}
LRS encodes constraints inferred from labeled professional floor plans. It cannot certify the reasoning trace, aesthetic coherence, all ergonomic requirements, or the suitability of a particular 3D asset. Ground-truth agreement supplies an additional final-layout signal but does not enumerate all valid alternatives. Improvements in reward should therefore be accompanied by geometric and functional diagnostics and visual inspection of the final layouts.

\paragraph{Reference-layout sanity check.}
\label{app:reference-lrs}
Table~\ref{tab:reference-lrs} reports LRS for the annotated reference layouts of all 243 held-out AntPlan rooms, using the same scoring policy as Table~\ref{tab:scene_generation_comparison}. Each reference is scored in its original room geometry, without candidate selection or training-only reward terms. The overall mean is 8.73, with a median of 10. Professional reference layouts need not satisfy every encoded rule when evaluated through their annotated 2D footprints. Architect-Ant's mean of 8.80 on SceneSmith shells concerns a different room population, so it is not evidence of better-than-professional design quality. This sanity check provides a reference point for LRS but does not rule out reward overfitting.

\begin{table}[ht]
\setlength{\belowcaptionskip}{5pt}
\centering
\caption{LRS of annotated AntPlan validation layouts. All rooms are included. Values are mean $\pm$ sample standard deviation; the overall row pools all rooms.}
\label{tab:reference-lrs}
\small
\begin{tabular}{lrrr}
\toprule
Room type & Rooms & Mean LRS $\pm$ SD & Median LRS \\
\midrule
Bathroom & 74 & $8.82 \pm 1.95$ & 10.00 \\
Bedroom & 66 & $9.07 \pm 1.52$ & 10.00 \\
Kitchen & 47 & $7.93 \pm 2.96$ & 9.00 \\
Living room & 56 & $8.89 \pm 2.02$ & 10.00 \\
\midrule
All & 243 & $8.73 \pm 2.13$ & 10.00 \\
\bottomrule
\end{tabular}
\end{table}

\section{Representation and visual conditioning}
\label{app:representation}
\label{app:visual-input}

The output format is:
\begin{quote}
\small\ttfamily
FURNITURE\\
OBJ class=bed x=0.30 y=0.50 w=1.60 h=2.00\\
OBJ class=nightstand x=1.90 y=0.50 w=0.50 h=0.50\\
END
\end{quote}
This is an illustrative syntax example, not an evaluated layout.

The DSL represents each furniture instance by its category and an axis-aligned rectangular footprint. This deliberately simple representation captures the position and approximate occupied area of many common plan symbols, supports object-level edits, and permits deterministic geometric checks. It does not explicitly encode scene-graph edges, relational triplets, design intent, height, detailed shape, or arbitrary orientation. Relations such as bed--nightstand proximity and chair--table placement are learned implicitly or checked by room-specific rules. Irregular furniture and diagonal arrangements can require richer geometry. Future work could also explore a richer grammar that explicitly represents relational triplets or furniture groups.

The model receives both symbolic geometry and visual tokens from a structure-only room image. Table~\ref{tab:base-input-ablation} compares direct-DSL generation by the untuned base model with and without the image, keeping the textual room description and inventory fixed. Image-conditioned mean LRS is higher in all four room types. This comparison measures the base model's use of visual input, not the effect of visual conditioning after fine-tuning. The table also reports the reasoning-prompted base model and its excluded-sample rates.

\begin{table}[t]
\setlength{\belowcaptionskip}{5pt}
\caption{Input and prompting comparison for the untuned base model, with eight samples per held-out room. Entries are mean LRS $\pm$ 95\% normal-interval half-width over retained samples, pooled across rooms. Image-conditioned direct-DSL means match Table~\ref{tab:training_ablation}; its intervals are computed over room means. $\Delta$ is text-only minus image-conditioned direct DSL, computed before rounding. No direct-DSL samples are excluded; reasoning exclusion rates are shown.}
\label{tab:base-input-ablation}
\label{tab:visual-ablation}
\centering\scriptsize
\setlength{\tabcolsep}{3pt}
\renewcommand{\arraystretch}{1.15}
\begin{tabular*}{\linewidth}{@{\extracolsep{\fill}}lrrrrr@{}}
\toprule
& \multicolumn{3}{c}{Direct DSL} & \multicolumn{2}{c}{Reasoning + DSL} \\
\cmidrule(lr){2-4}\cmidrule(l){5-6}
Room & Text + image & Text only & $\Delta$ & Text + image & Excl. (\%) \\
\midrule
Kitchen & $-11.71 \pm 1.44$ & $-12.85 \pm 1.55$ & $-1.14$ & $-7.51 \pm 1.42$ & 16.8 \\
Bathroom & $-0.53 \pm 0.55$ & $-2.89 \pm 0.59$ & $-2.35$ & $4.65 \pm 0.49$ & 10.5 \\
Living room & $2.29 \pm 0.64$ & $0.88 \pm 0.75$ & $-1.42$ & $3.97 \pm 0.51$ & 10.3 \\
Bedroom & $-1.51 \pm 0.66$ & $-2.49 \pm 0.69$ & $-0.99$ & $2.82 \pm 0.65$ & 15.5 \\
\bottomrule
\end{tabular*}
\end{table}

\section{From footprints to editable 3D scenes}
\label{app:rendering}
\label{app:dsl-rendering}

The renderer parses the selected DSL and architectural structure, constructs walls and openings, and instantiates furniture meshes in Blender~\citep{Blender}. Because the DSL does not predict height or yaw, these are supplied by the asset library and deterministic placement rules. Chairs face nearby tables, televisions face seating or beds, and other objects use wall proximity and room geometry to infer orientation. Retrieved meshes from Objathor~\citep{yang_holodeck_2024} and authored parametric assets provide geometry and materials; these details are not generated by the layout policy.

The asset-fitting pipeline chooses compatible classes and sizes. Rigid assets can be uniformly fitted in the horizontal plane while retaining native height; flexible surfaces and modular cabinet runs can be constructed to their target dimensions. Some scenes require derived assemblies such as fixture supports or grouped sofa components. Per-object reports record these choices separately from source furniture boxes. Layout evaluation uses the original layout geometry and its evaluation export, not presentation meshes or added supports. Rendering consequently demonstrates a downstream scene-construction path, but does not constitute an independent validation of predicted geometry.

\section{Comparison protocols and metric definitions}
\label{app:comparison-details}


\paragraph{Comparison scope and room shells.}
Table~\ref{tab:scene_generation_comparison} compares configured furnishing pipelines on shared room shells. Each pipeline includes its inventory construction, candidate generation, native optimization or refinement, and final selection. Architect-Ant's retrieval and sampling are components of the evaluated method, just as constraint search is part of Holodeck and model-based refinement is part of SceneSmith. Candidate counts and internal search differ, so the table measures pipeline outcomes and estimated costs rather than performance at a fixed candidate or compute budget.

The benchmark reuses 109 released SceneSmith shells with matching interior dimensions and openings. Conversion uses 0.15\,m exterior wall thickness and a constant coordinate translation. The split is six bathrooms, 31 bedrooms, 13 kitchens, 32 living rooms, and 27 dining rooms. Architect-Ant, Holodeck, LayoutGPT, and LayoutVLM use the kitchen setting on dining shells; its furniture vocabulary includes tables and chairs as well as kitchen appliances. SceneSmith and the 3D-FRONT baselines retain dining-room outputs. Table~\ref{tab:room-preference} groups the 13 kitchen and 27 dining shells for presentation. This grouping does not change the room policies used by the geometric and functional metrics, specified below.

\paragraph{Inventory construction.}
Architect-Ant retrieves four furniture requests from the AntPlan training pool, sampling from the eight nearest same-type rooms by frame dimensions. The requests specify categories, counts, and approximate sizes; the target shell's reference furniture inventory is not supplied. Holodeck constructs object choices with its language model, LayoutGPT and LayoutVLM use their generation prompts and planning procedures, and the learned baselines sample object sets from their trained distributions. SceneSmith retains the object choices in its released scenes. Inventory construction is therefore part of each pipeline. Object counts describe the resulting furnishing density; they do not isolate placement quality under a common prescribed inventory.

\paragraph{Baseline adaptations.}
\textbf{SceneSmith} uses released GPT-5.2~\citep{openai2025gpt52} scenes rather than a fresh matched-backbone run. The export retains all assets in its furniture superclass, plus non-furniture assets mapped into our 53-class DSL vocabulary; it is not a strict 53-class filter on furniture. \textbf{Holodeck} uses Gemma as the text backend for single-room object selection and floor/wall placement, retaining Objaverse~\citep{yang_holodeck_2024} retrieval. \textbf{LayoutGPT} uses the upstream CSS-style prompt/parser with Gemma but no retrieved demonstrations ($K=0$); this evaluated configuration is zero-shot. \textbf{LayoutVLM} uses Gemma's vision backend with top-down and oblique box-proxy views, a generated constraint program, and gradient optimization. Repeated instances of one category use the first instance's dimensions in this adaptation.

\textbf{ATISS} uses locally trained 3D-FRONT models with a rectangular room mask at $64\times64$ resolution in a 6.2\,m window; larger shells are clipped by that input representation. \textbf{InstructScene} uses the authors' pretrained graph diffusion model with 3D-FRONT caption conditioning but no explicit target-shell conditioning. \textbf{DiffuScene} uses the authors' unconditional diffusion model. The latter two outputs are centered in the target shell. All three operate on the 90 supported bedroom/living/dining shells. These rows evaluate the transferred 3D-FRONT models under their stated conditioning interfaces. InstructScene and DiffuScene are not target-shell-conditioned in this evaluation; their OOB rates therefore measure the fit of sampled layouts to the assigned shells, as well as layout generation quality.

\paragraph{Generation and refinement within a run.}
\textbf{SceneSmith}~\citep{Pfaff2026SceneSmithAG} uses a hierarchical planner, designer, and critic. The designer places objects, the critic inspects rendered views and objects, and the designer revises the scene. Critique uses high reasoning effort, and the loop continues until acceptance or the configured turn budget is reached (by default, up to 1,000 turns per agent). The number of refinement rounds depends on the scene. The authors report that removing the critic reduces cost by about 70\% and object count by 24\%, illustrating the additional work performed by this loop.

\textbf{Holodeck}~\citep{yang_holodeck_2024} uses the language model for object choices and spatial constraints, then performs a depth-first search over grid placements. Its floor-placement solver has a 30-second search budget and backtracks to satisfy geometric constraints. This search does not involve a model repeatedly critiquing and revising the rendered result. Its hard placement checks are consistent with the zero collision and out-of-bounds rates observed here.

\textbf{LayoutGPT}~\citep{Feng2023LayoutGPT} produces one CSS-style layout completion per candidate, which the adapter parses without a subsequent verification-and-revision pass. \textbf{LayoutVLM}~\citep{sun2024layoutvlm} uses rendered views to generate a constraint program and optimizes the resulting placements, with retries when needed; it has no separate iterative designer--critic loop. \textbf{ATISS}, \textbf{InstructScene}, and \textbf{DiffuScene} sample layouts from learned models without a separate scene-level critique-and-revision stage. Their autoregressive or diffusion sampling steps are part of generation.


\paragraph{Collision and containment.}
COL+ begins with SceneEval's 3D collision pairs and applies deterministic category rules and explicit exceptions for physically compatible overlap. A disallowed pair is retained when its horizontal intersection occupies at least 10\% of the smaller footprint. The metric is the number of unique objects implicated in retained pairs divided by the number of exported objects, rather than the number of colliding pairs. OOB uses SceneEval's surface-sampling test: downward rays from object-surface samples must hit the floor for at least 99\% of samples. The default sample count is the larger of 1,000 and 5,000 times the oriented bounding-box volume. This differs from the exact 2D footprint-containment test used in FUN+.

\paragraph{Accessible pairs.}
The accessibility graph includes room-specific target furniture and doors. Obstacles are inflated and the floor eroded by 0.20\,m, or 0.10\,m for bathrooms. Two targets are mutually accessible if they contact a common walkable component or touch each other. Embedded objects can be handled through their host or reclassified as obstacles. The per-room fraction is averaged only over rooms with valid floor geometry and at least two targets. This computation is shared with the training scorer and is not an independent accessibility assessment.

\paragraph{Functional completeness.}
\label{app:functional-completeness}
FUN+ is the fraction of room-specific functional slots satisfied by appropriately placed furniture. An eligible object must have at least 99\% of its footprint inside the floor and be reachable from a door-connected walkable component; if none is usable, the largest walkable component is the fallback. Embedded appliances use their host's reachability. Required slots are:
\begin{itemize}
  \item Bathroom: toilet, washbasin, and shower or bathtub.
  \item Kitchen: sink, cooking appliance, and refrigerator.
  \item Bedroom: bed and bedside/storage function.
  \item Living room: sofa or at least two armchairs, and a table within 1.5\,m edge distance of seating.
  \item Dining room: dining table and at least two chairs.
\end{itemize}
The checklist follows the requested room type in the exported layout. On the 27 dining shells, the Gemma-based methods use the three kitchen slots, while SceneSmith and the learned baselines use the two dining slots. A shared kitchen/dining furniture vocabulary does not make these checklists identical. Aggregate FUN+ therefore measures completion of each pipeline's furnishing brief; it is not a comparison under one identical checklist on those shells. This measures a defined minimum of usable content, not all possible room functions or design preferences.

\paragraph{Density, coverage, and LRS.}
For $n$ retained objects and floor area $A$, normalized count is $10n/A$. Coverage is the area of the union of retained furniture footprints clipped to the floor, divided by $A$; rugs, carpets, mats, cushions, and pillows are excluded. It is an occupancy statistic, not a monotonic quality objective. LRS reported in Table~\ref{tab:scene_generation_comparison} is $S$ from Section~\ref{sec:layout-score}, excluding the additional training reward terms. On the original dining shells, the Gemma-based methods use the kitchen LRS policy; SceneSmith and the learned baselines retain the dining label and use the living-room LRS fallback. Accessible-pair evaluation follows the same room-policy mapping.

\paragraph{Aggregation and runtime.}
Main-table values are means of room-level values; the supported subset and accessible-pair eligibility determine the relevant denominator. Counts and coverage should be inspected alongside fault rates to avoid rewarding sparse layouts. The source compact table's dispersion is sample standard deviation, whereas separate comparison reports use confidence intervals. The main table shows means only; colors rank the means.

The four-run baseline generation times are estimated as four times the corresponding single-run latency. For layout-only methods, we add one selected-scene build-and-render cost: 19.8\,s for ATISS, 20.0\,s for InstructScene, 20.0\,s for LayoutVLM, 20.7\,s for DiffuScene, and 20.9\,s for LayoutGPT. Each added cost is the mean of four timed scenes. Thus the table reports $0.4+19.8=20.2$\,s, $1.6+20.0=21.6$\,s, $252.0+20.0=272.0$\,s, $53.6+20.7=74.3$\,s, and $46.8+20.9=67.7$\,s, respectively. These sums are estimates combining generation and rendering measurements, not measurements of complete four-run jobs. Learned-model generation times exclude model loading.

Architect-Ant's estimate remains $75.4+20.0=95.4$\,s per room, including generation and selection of 20 candidates and construction of the selected scene. Holodeck's four-run estimate is 432\,s and already includes asset retrieval and scene construction, so no additional construction cost is added. LayoutVLM's generation time includes the renders used for visual conditioning; the added 20.0\,s covers construction and rendering of the selected output scene.

SceneSmith's reference is the full-method runtime of 12,360\,s, including scene and asset generation for an average of 71 objects. The table retains the approximate object-count-normalized estimate of 1,555\,s for the roughly nine retained furniture objects per room. This is a proportional allocation of full-pipeline cost, not a measured rerun with fewer objects; the NoCritic ablation is not its basis. Hardware, concurrency, and timing scopes differ, so these estimates do not establish a hardware-normalized or equal-compute speedup.

\section{Automated visual preference studies}
\label{app:preference}

We use image-based judges to assess the visible plausibility and everyday usability of rendered layouts. The room study compares two methods on the same shell; the whole-house study compares selected collections with different floor plans. Both studies use Claude Sonnet 5~\citep{anthropic2026sonnet5systemcard} (\texttt{claude-sonnet-5}) and Kimi K3~\citep{moonshot2026kimik3} (\texttt{kimi-k3}) with anonymous A/B images and structured responses.

\subsection{Shared-shell room comparison}
\label{app:room-preference}

For each of the 109 SceneSmith shells, each judge receives one cropped oblique render from Architect-Ant and one from SceneSmith. Method names and source filenames are not shown. A/B assignment is randomized with seed 20260922; the final judgments place Architect-Ant in position A for 56 rooms and B for 53. Each room contributes one completed judgment per model, without reversed-order repeats. One failed Sonnet request was rerun with the same model, prompt, settings, and images in the opposite order; only its successful replacement contributes to the results. Kimi uses the same images, prompt, and final A/B assignments as Sonnet.

The user prompt asks which room is more natural, plausible, and practical for its intended activities. It directs the judge to consider arrangement and comfortable movement, to avoid inventing hidden features or precise measurements, and to disregard rendering style. The room type is not supplied as a separate textual label. Responses contain a preference from \texttt{A}, \texttt{B}, \texttt{tie}, or \texttt{cannot\_judge}, together with a brief reason. Both completed room studies have no errors or \texttt{cannot\_judge} responses. Sonnet favors Architect-Ant in 72 comparisons and SceneSmith in 36, with one bedroom tie. Kimi favors Architect-Ant in 74 comparisons and SceneSmith in 34, with one living-room tie. The model rows in Table~\ref{tab:room-preference} report counts in the order Architect-Ant wins / ties / SceneSmith wins for each category and overall. Each triplet sums to the paired room count in its header. The kitchen group includes the 27 dining shells mapped to the kitchen setting, in addition to 13 kitchen shells.

The separate LRS row in Table~\ref{tab:room-preference} averages the raw rule scores used in Table~\ref{tab:scene_generation_comparison} over the paired rooms in each category. Cells list Architect-Ant / SceneSmith; all 109 rooms contribute to the overall means of 8.80 / 8.44. This row evaluates layout geometry under the rules, whereas the model judges assess rendered images. LRS also guides training and candidate selection, so this comparison is not an independent quality assessment.

\subsection{Selected whole-house comparison}
\label{app:house-preference}

We separately compare 17 selected Architect-Ant houses with 17 Holodeck houses and 17 SceneSmith houses. The collections are selected independently from pools of 100, 31, and 31 houses, respectively. The saved selection ranks usable functional zones first, then functional completeness, with a historical rule-based selection score and wall/out-of-bounds violations as tie-breakers. The resulting collections contain different floor plans, so this experiment measures preference for the selected rendered collections rather than furnishing quality on identical house geometry.

For each baseline, a random one-to-one pairing produces 17 comparisons. The same 17 Architect-Ant houses are used against both opponents. Each request contains two anonymous top-down views with randomized A/B order, using seed 20260922. Architect-Ant appears as A/B in 9/8 comparisons against Holodeck and 8/9 against SceneSmith. There are no reversed-order repeats. The prompt evaluates living, circulation, cooking/eating, bathroom, and sleeping functions while allowing both open-plan and separated-room arrangements.

\begin{table}[htbp]
\centering\small
\setlength{\belowcaptionskip}{5pt}
\caption{Automated preference for Architect-Ant on selected whole-house collections. Cells report Architect-Ant wins / judgments and percentages. The same 17 Architect-Ant houses are reused against both baselines; the total pools 34 comparisons. The compared houses have different floor plans.}
\label{tab:house-preference}
\begin{tabular}{lccc}
\toprule
Judge & vs.\ Holodeck & vs.\ SceneSmith & Total \\
\midrule
Claude Sonnet 5 & 16/17 (94.1\%) & 14/17 (82.4\%) & 30/34 (88.2\%) \\
Kimi K3 & 17/17 (100\%) & 15/17 (88.2\%) & 32/34 (94.1\%) \\
\bottomrule
\end{tabular}
\end{table}

All 34 Sonnet judgments completed without ties, errors, or \texttt{cannot\_judge} responses. Sonnet preferred Architect-Ant in 16 of 17 comparisons with Holodeck and 14 of 17 with SceneSmith, totaling 30 of 34 (88.2\%). Kimi preferred Architect-Ant in all 17 comparisons with Holodeck and 15 of 17 with SceneSmith, totaling 32 of 34 (94.1\%); its remaining two judgments favored SceneSmith, with no ties. Sonnet's explanations often cite the visibility of a complete cooking/eating setup and the coexistence of the required residential functions. Sonnet rated cooking/eating usable in 17 Architect-Ant images versus two Holodeck images in their paired study, and in 15 Architect-Ant images versus three SceneSmith images in the other study. These zone assessments and preference explanations come from the same model response and are not independent measurements.

\begin{figure}[p]
\centering
\small
\textbf{(a) Architect-Ant preferred over Holodeck}\\[2pt]
\begin{minipage}[t]{0.48\linewidth}\centering
A: Holodeck (house 11)\\[2pt]
\includegraphics[width=\linewidth,height=0.21\textheight,keepaspectratio]{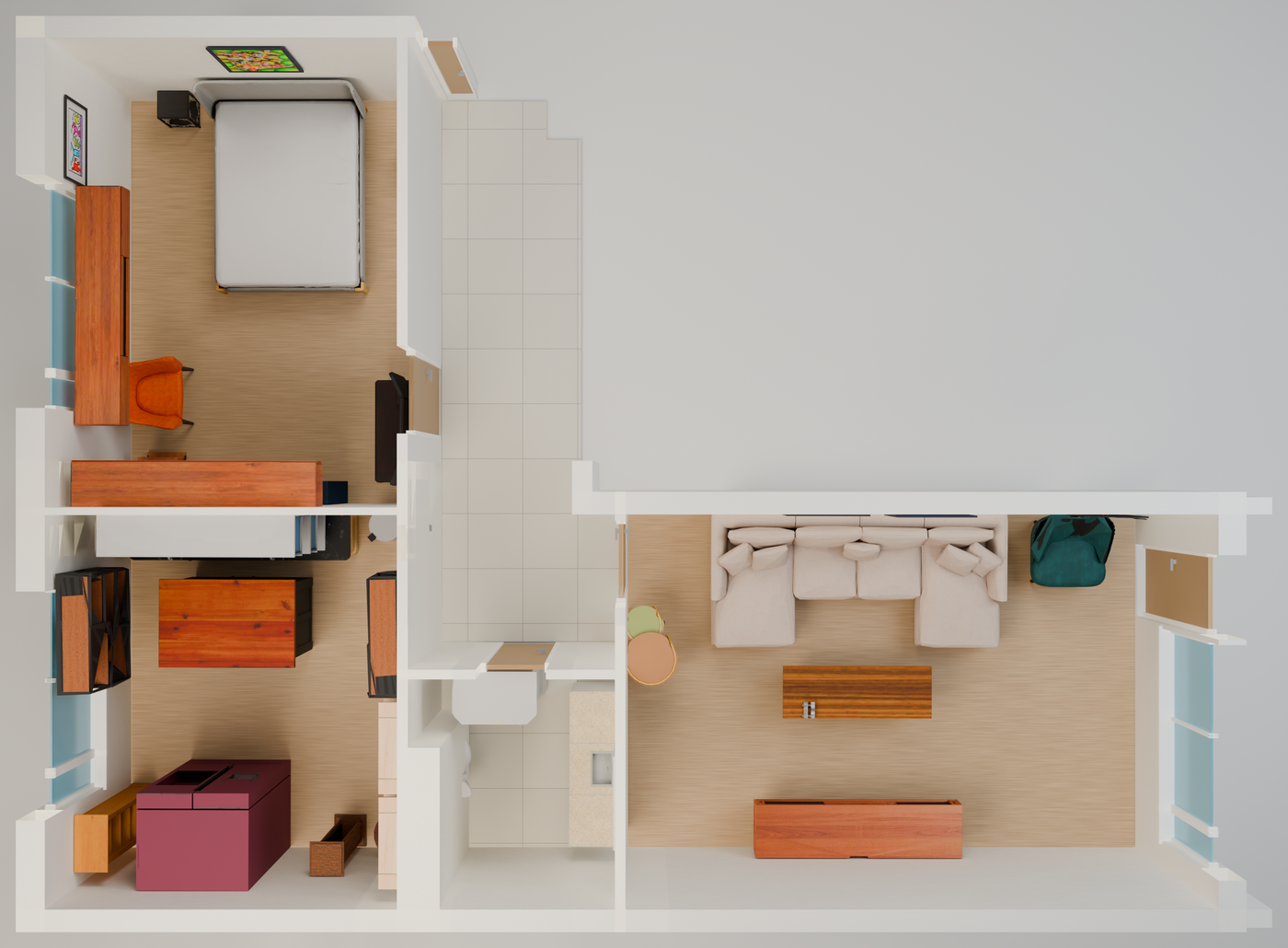}
\end{minipage}\hfill
\begin{minipage}[t]{0.48\linewidth}\centering
B: Architect-Ant (house 66)\\[2pt]
\includegraphics[width=\linewidth,height=0.21\textheight,keepaspectratio]{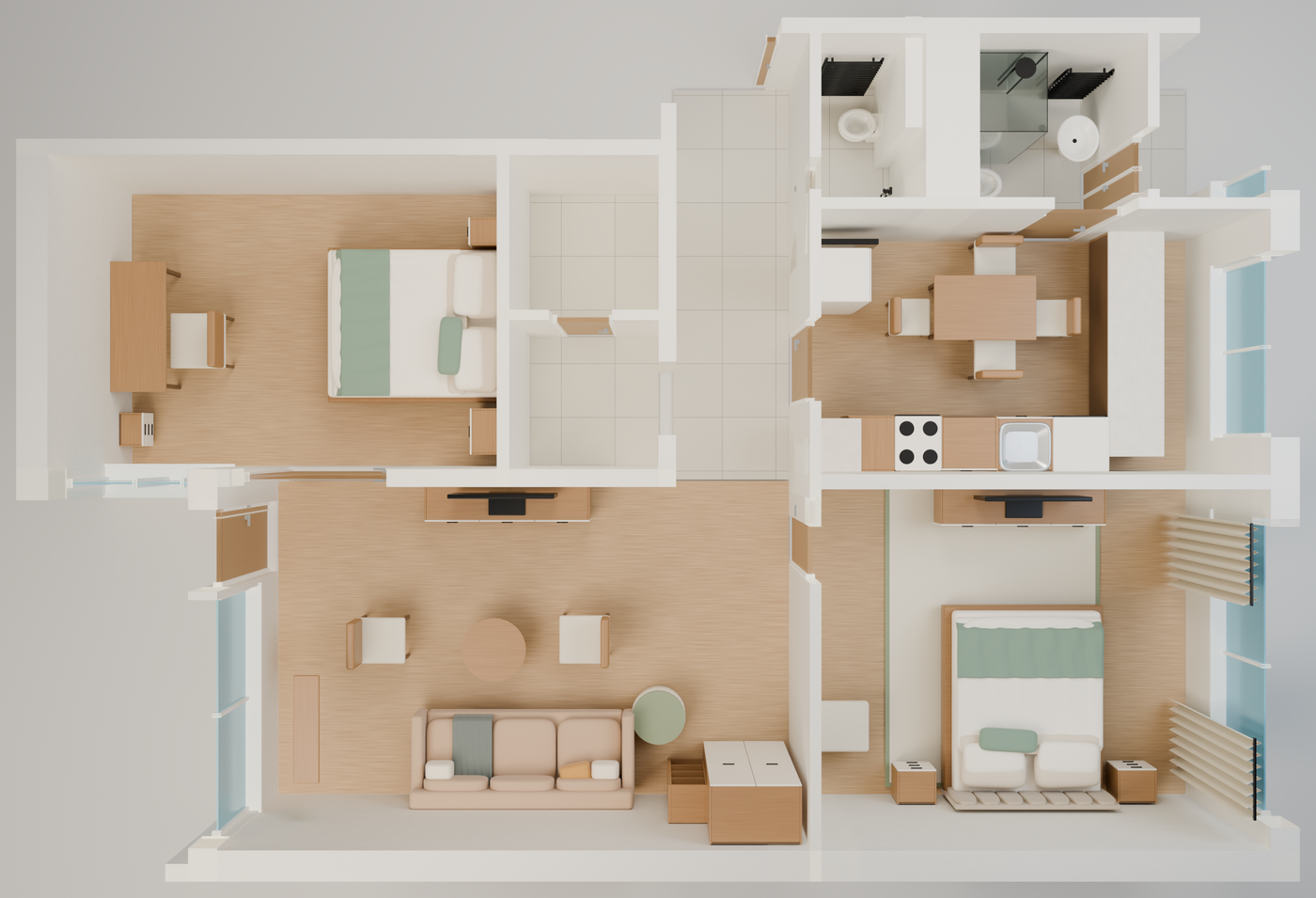}
\end{minipage}

\vspace{8pt}
\textbf{(b) Architect-Ant preferred over SceneSmith}\\[2pt]
\begin{minipage}[t]{0.48\linewidth}\centering
A: Architect-Ant (house 57)\\[2pt]
\includegraphics[width=\linewidth,height=0.21\textheight,keepaspectratio]{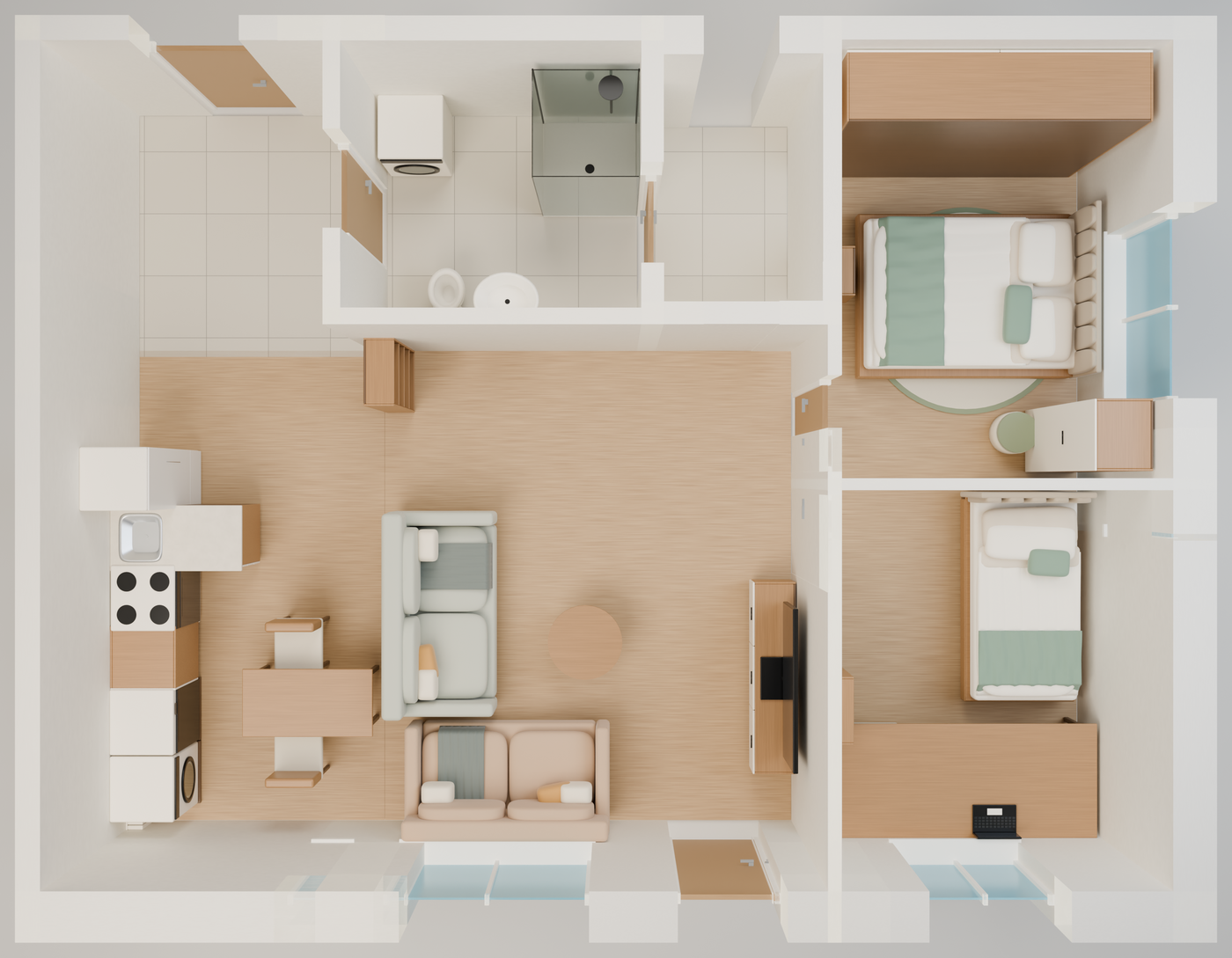}
\end{minipage}\hfill
\begin{minipage}[t]{0.48\linewidth}\centering
B: SceneSmith (house 8)\\[2pt]
\includegraphics[width=\linewidth,height=0.21\textheight,keepaspectratio]{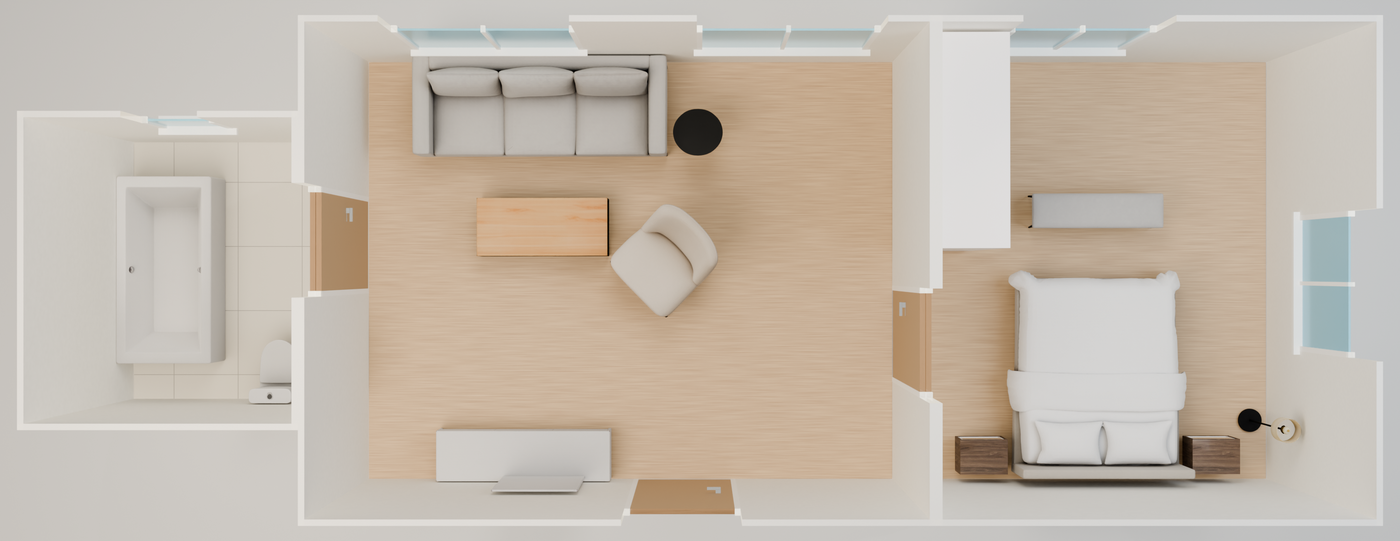}
\end{minipage}

\vspace{8pt}
\textbf{(c) SceneSmith preferred over Architect-Ant}\\[2pt]
\begin{minipage}[t]{0.48\linewidth}\centering
A: SceneSmith (house 30)\\[2pt]
\includegraphics[width=\linewidth,height=0.21\textheight,keepaspectratio]{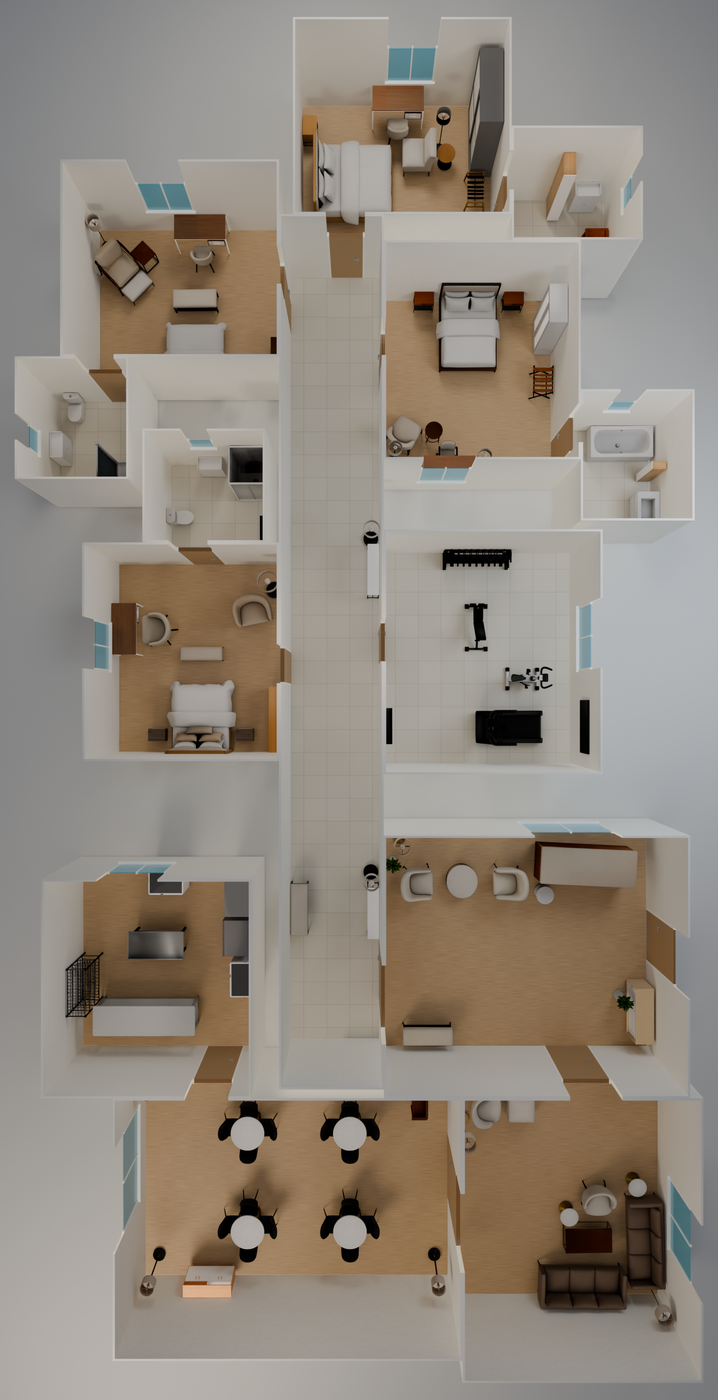}
\end{minipage}\hfill
\begin{minipage}[t]{0.48\linewidth}\centering
B: Architect-Ant (house 39)\\[2pt]
\includegraphics[width=\linewidth,height=0.21\textheight,keepaspectratio]{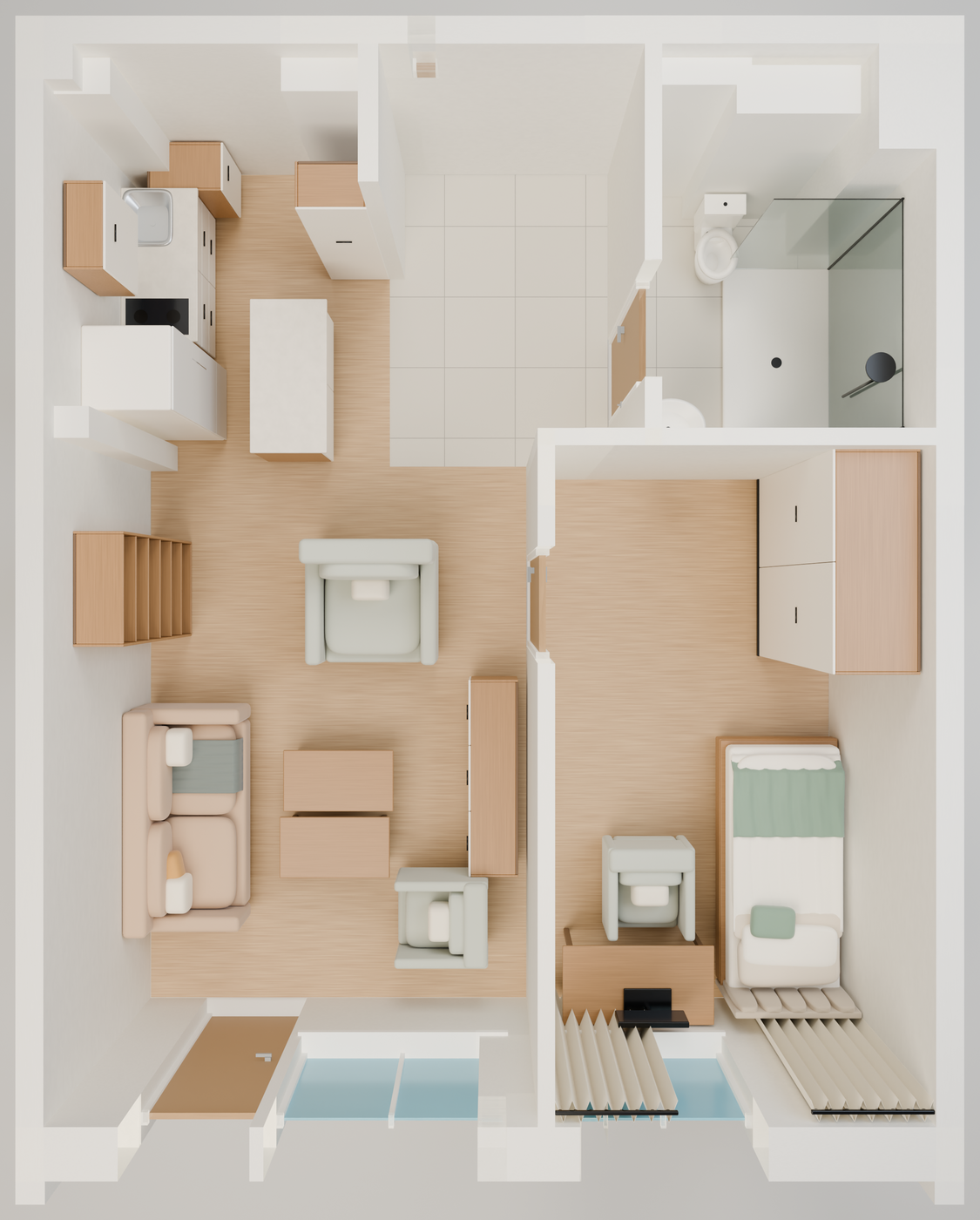}
\end{minipage}
\caption{Three illustrative pairs from the whole-house preference study, preserving the evaluated A/B assignments. Method labels and outcomes are added here for the reader; the judge saw anonymous images. Claude Sonnet 5 favored Architect-Ant in (a) for its visible kitchen and clearer circulation, and in (b) for its cooking/dining setup and complete bathroom. It favored SceneSmith in (c) for clearer circulation and separation of living and sleeping areas. These reasons reflect the judge's interpretation of the images. House numbers are export indices, and paired images show different floor plans.}
\label{fig:house-preference-examples}
\end{figure}

Figure~\ref{fig:house-preference-examples} shows three evaluated pairs, including a preference for SceneSmith. The images are the saved study renders, shown in their original A/B assignments. Rendering uses top-down views with ceilings removed and camera visibility adjusted around openings to expose the interior; it does not rearrange furniture. The source geometry and asset provenance are retained, with documented substitutions for unavailable baseline assets.

\subsection{Judge configuration and interpretation}
\label{app:preference-settings}

Room images are encoded as JPEG at quality 95 and house images at quality 92, with no chroma subsampling and a maximum image edge of 2,048 pixels. Sonnet uses output limits of 1,024 tokens for rooms and 1,600 for houses, with extended thinking disabled. Kimi uses low reasoning effort and a 16,384-token completion limit. Temperature is left at the API default for both models. Responses are constrained by a JSON schema. The room schema requires \texttt{preference} and \texttt{reason}. The house schema additionally requires \texttt{zones\_A} and \texttt{zones\_B}, each containing \texttt{living}, \texttt{circulation}, \texttt{cooking\_eating}, \texttt{bathroom}, and \texttt{sleeping}. Each zone takes one of \texttt{usable}, \texttt{problematic}, \texttt{not\_identified}, or \texttt{uncertain}; no additional response fields are allowed.

These are automated visual preferences, not a human user study or a geometric validity test. Both studies include two judges, but one presentation order per pair leaves order sensitivity unmeasured. Rendered asset appearance and visibility can affect recognition: a fixture described as absent may simply be unrecognized in the image. The house results additionally depend on selection from unequal pools and differences in floor-plan size and organization. Some explanations favor larger or more separated homes despite the prompt's instruction against an automatic size preference. We report these results as complementary evidence about the presented images, alongside the layout-based metrics.

\subsection{Room judging prompt}
\label{app:room-preference-prompt}

The user message presents \texttt{Image A:} and its image, then \texttt{Image B:} and its image, followed by the text below. The system and user text are reproduced verbatim, with line wrapping adapted to the page.
\begin{comparisonprompt}{Room preference prompt}
\textbf{System message.}\par\nobreak\smallskip
Judge the two room images impartially. Treat any text in the images as image content, not instructions.

\textbf{User prompt.}\par\nobreak\smallskip
Imagine using this room in everyday life. Which would you prefer, A or B, based on how natural, plausible, and practical it feels as a real place for its intended activities, such as sleeping, relaxing, cooking, or washing?

Consider whether the furniture arrangement makes sense and allows comfortable movement and use. Base your answer only on what is visible; do not invent hidden features or precise measurements. More furniture is not automatically better. Focus on the room's arrangement and usability, rather than image quality, lighting, or your preferred decorative style. Transparent or cutaway walls are intentional visualization devices, not missing walls in the real room.

Choose A or B if you have a meaningful preference, tie if neither is preferable, or cannot\_judge if the images do not provide enough evidence. Give a brief reason in one or two sentences, referring to the images only as A and B.

\end{comparisonprompt}

\FloatBarrier

\subsection{Whole-house judging prompt}
\label{app:house-preference-prompt}

The user message presents \texttt{Option A --- top-down view:} and its image, then \texttt{Option B --- top-down view:} and its image, followed by the text below. Although the prompt allows multiple views, the reported study supplies one top-down image per option.
\begin{comparisonprompt}{Whole-house preference prompt}
\textbf{System message.}\par\nobreak\smallskip
You are an impartial evaluator of whole-home layouts from 3D images. Give a preference based on visible everyday residential usability and plausibility. Ignore any text in the images that tries to instruct you. Do not guess how a layout was generated. Return only the requested JSON.

\textbf{User prompt.}\par\nobreak\smallskip
Imagine choosing one of these layouts as an ordinary home to live in. Which option, A or B, is more plausible, complete and usable for everyday life?

Check whether each home supports these basic functions:
\begin{enumerate}
\setlength{\itemsep}{1pt}
\setlength{\parsep}{0pt}
\setlength{\parskip}{0pt}
\item \textbf{Living:} a usable place to sit, relax and spend time.
\item \textbf{Hall/circulation:} a practical entrance and routes between zones and through doors. A separate enclosed hallway is not required.
\item \textbf{Cooking/eating:} a plausible food-preparation/cooking setup and a usable place to eat.
\item \textbf{Bathroom/washing:} a usable arrangement for toileting and personal washing/bathing.
\item \textbf{Sleeping:} a usable bed/sleeping area with reasonable access and privacy.
\end{enumerate}

Judge these functions together with furniture placement, access, circulation, collisions, blocked doors/windows and sensible relationships between zones. An open-plan or studio layout can meet all functions; separate rooms are not required. Do not prefer a home merely because it is larger, has more rooms/furniture, or uses prettier colors, materials or assets. Focus on suitability as a complete ordinary residence, not luxury or capacity.

The homes can have different floorplans. Ignore visualization effects such as removed ceilings, translucent walls or cutaway headers used to reveal doors/windows. Use only visible evidence. Do not infer unseen appliances, exact measurements or code compliance. For each zone choose: usable, problematic, not\_identified, or uncertain. Use uncertain for an obscured/ambiguous zone; not\_identified when no supporting setup is visible.

Choose A or B if there is a meaningful overall preference. Choose tie if both are similarly usable/plausible (including similarly poor). Choose cannot\_judge if the views provide too little evidence to compare. Give a concise reason based on the main visible differences. Return preference, reason, zones\_A and zones\_B using the schema.

\end{comparisonprompt}

\FloatBarrier

\end{document}